\documentclass[acmsmall,screen]{acmart}

\acmJournal{PACMPL}
\acmVolume{1}
\acmNumber{POPL} 
\acmArticle{1}
\acmYear{2022}
\acmMonth{1}
\acmDOI{} 
\startPage{1}

\setcopyright{none}

\usepackage{booktabs}   
\usepackage{subcaption} 
\newtheorem{definition}{Definition}[section]
\newtheorem{problem}{Problem}[section]
\DeclareMathOperator*{\minimize}{minimize}

\newtheorem{remark}{Remark}[section]
\usepackage{algorithm}
\usepackage[noend]{algpseudocode}
\newcommand{\comm}[1]{\textcolor{blue}{#1}}

\begin{document}

\title[]{Neural Network Repair with Reachability Analysis}         


\author{Xiaodong Yang}
\affiliation{
  \department{Electrical Engineering and Computer Science}     
  \institution{Vanderbilt University}            
  \city{Nashville}
  \state{TN}
  \country{USA}                    
}
\email{xiaodong.yang@vanderbilt.edu}          

\author{Tom Yamaguchi}        
\affiliation{
  \institution{TRINA, Toyota NA R\&D}           
  \city{Ann Arbor}
  \state{MI}
  \country{USA}                   
}
\email{tomoya.yamaguchi@toyota.com}         

\author{Hoang-Dung Tran}                                   
\affiliation{
  \institution{University of Nebraska}           
  \city{Lincoln}
  \state{NE}
  \country{USA}                   
}
\email{trhoangdung@gmail.com}         

\author{Bardh Hoxha}        
\affiliation{
  \institution{TRINA, Toyota NA R\&D}           
  \city{Ann Arbor}
  \state{MI}
  \country{USA}                   
}
\email{bardh.hoxha@toyota.com}  

\author{Taylor T Johnson}
\affiliation{
  \department{Electrical Engineering and Computer Science}     
  \institution{Vanderbilt University}            
  \city{Nashville}
  \state{TN}
  \country{USA}                    
}
\email{taylor.johnson@vanderbilt.edu}    

\author{Danil Prokhorov}        
\affiliation{
  \institution{TRINA, Toyota NA R\&D}           
  \city{Ann Arbor}
  \state{MI}
  \country{USA}                   
}
\email{danil.prokhorov@toyota.com}   

\begin{abstract}


Safety is a critical concern for the next generation of autonomy that is likely to rely heavily on deep neural networks for perception and control. Formally verifying the safety and robustness of well-trained DNNs and learning-enabled cyber-physical systems (Le-CPS) under adversarial attacks, model uncertainties, and sensing errors is essential for safe autonomy. This research proposes a framework to repair unsafe DNNs in safety-critical systems with reachability analysis. 
The repair process is inspired by adversarial training which has demonstrated high effectiveness in improving the safety and robustness of DNNs. Different from traditional adversarial training approaches where adversarial examples are utilized from random attacks and may not be representative of all unsafe behaviors, our repair process uses reachability analysis to compute the exact unsafe regions and identify sufficiently representative examples to enhance the efficacy and efficiency of the adversarial training.

The performance of our repair framework is evaluated on two types of benchmarks without safe models as references. One is a DNN controller for aircraft collision avoidance with access to training data. The other is a rocket lander where our framework can be seamlessly integrated with the well-known deep deterministic policy gradient (DDPG) reinforcement learning algorithm. The experimental results show that our framework can successfully repair all instances on multiple safety specifications with negligible performance degradation. In addition, to increase the computational and memory efficiency of the reachability analysis algorithm in the framework, we propose a depth-first-search algorithm that combines an existing exact analysis method with an over-approximation approach based on a new set representation. Experimental results show that our method achieves a five-fold improvement in runtime and a two-fold improvement in memory usage compared to exact analysis. 

\end{abstract}




\renewcommand\footnotetextcopyrightpermission[1]{} 
\fancyfoot{}

\maketitle
\thispagestyle{empty}

\section{Introduction}
Despite success of deep neural networks (DNNs) in various applications, trustworthiness is still one of the main issues preventing widespread use. 
Research has shown that DNNs may generate undesired behaviors even with the slightest perturbations on input data.
Recently, many techniques for analyzing behaviors of DNNs have been presented ~\cite{katz2019marabou,tran2020nnv,dutta2018output,tran2019fm,singh2019abstract,xiaodong21,yang2021reachability,yang2020reachability,botoeva2020efficient,Urban20, urban2021review, frankle2020pruning,sotoudeh2021syrenn,xiong2021scalable, anderson2020neurosymbolic, wang2020abstract}. Given a DNN, these works can generate a safety certificate over an input-output specification \cite{seshia2018formal}. However, due to the black-box nature of DNNs, training safe DNNs or repairing their erroneous behaviors remains a challenge.







Existing works to improve the safety and robustness of DNNs can be classified into two main categories. The first category relies on singular adversarial inputs to make specialized modifications on neural weights that likely result in misbehavior. 
In~\cite{sohn2019search}, the authors propose a technique named \textit{Arachne}. There, given a set of finite adversarial inputs that cause undesired behaviors, with guidance of a fitness function, \textit{Arachne} searches and subsequently modifies neural weights that are likely related to these undesired behaviors. The method supports specifications consisting of a finite set of inputs instead of continuous regions. 
In~\cite{goldberger2020minimal}, the authors propose a DNN verification-based method that modifies undesirable behavior of DNNs by manipulating neural weights of the output layer. The correctness of the repaired DNN can be formally proved with the verification technique. However, the repair process is limited to a single adversarial example in each iteration. Typically, DNNs may contain multiple unsafe input regions over a continuous domain. In addition, the approach relies on modifications of the output layer, which may limit its capability.
The second category utilizes adversarial examples for retraining. 
Adversarial training works such as~\cite{goodfellow2014explaining, madry2017towards} have demonstrated that incorporating adversarial examples into the training process can improve the robustness of DNNs. However, DNNs may misbehave over continuous regions and infinitely many adversarial examples. Despite the robustness improvements, this training approach cannot guarantee safety for the learned DNNs. To solve this issue, some researchers incorporate reachability analysis in this process, such that they can train a model that is provably safe against norm-bounded adversarial attacks~\cite{wong2018provable, mirman2018differentiable}. Given a norm-bounded input range, these approaches over approximate the output reachable domain of DNNs with convex regions. Then they conduct robust optimization by minimizing the worst-case loss over these regions, which aims to migrate all the outputs to a desired domain. The primary issue of these approaches is that the approximation error accumulates during computation. For large input domains or complex DNNs, their approximated reachable domain can be so conservative that a low-fidelity worst-case loss may result in significant accuracy degradation.  
One promising alternative is to utilize exact reachability analysis methods~\cite{tran2020nnv,xiaodong21,tran2019fm, bak2020cav, tran2020cav}. 
These methods can compute the exact reachable set of DNNs and identify all the unsafe input regions.



In this paper, we propose a framework to repair DNNs, which combines adversarial training with exact reachability analysis of DNNs. We demonstrate the method and repair DNN controllers with respect to input-output safety specifications. In each iteration of the process, unsafe input regions are computed and incorporated into the training data. The iterative process will terminate once a model candidate is verified safe and also its performance is above a threshold.
At the heart of our approach, we utilize a novel exact reachability method that is optimized for identification of the unsafe input regions.
We also integrate our framework with learning algorithms of DNNs, specifically the deep reinforcement learning (DRL). 
The feasibility and effectiveness of this DNN repair framework is demonstrated on two types of benchmarks. 
One is an unsafe DNN of a horizontal collision avoidance system where the training data is accessible. 
For unsafe DNNs where the training data is available, the repair algorithm merges the unsafe regions of the model candidate to the training data in each loop, as shown in Figure~\ref{fig:framework}. 
The other benchmark is an application of our framework on a DNN trained through a DRL algorithm. 
Here, the repair algorithm will be slightly modified since DRL is utilized to learn policies that maximize the expectation of rewards in the long term, as well as ensure reasonable behaviors by avoiding violations of safety constraints. 
The risk in DRL is normally associated with the inherent uncertainty of the environment and the facet that even an optimal agent may perform unsuccessfully with such stochastic natures.  It is because the maximization of long-term rewards only involve finite environment and agent states, and it does not necessarily prevent rare occurrence of states that incur unsafe unsafe actions and subsequent safety violations. 
There is significant recent work on safe RL~\cite{islam_repairing_2020,xiong_robustness_2020,sohn_search_2019,deshmukh_runtime-safety-guided_2020,alshiekh_safe_2017,fulton_safe_nodate,bouton_reinforcement_2019,cheng_end--end_2019, huang_survey_2020}. 
Most existing work relies on high-fidelity knowledge of the environment dynamics and, to our best knowledge, there exist few approaches that can compute and eliminate risks from the environment uncertainty due to their non-determinism. In contrast, the advantages of our framework for DRL are threefold. Firstly, our framework can construct the all possible states
as well as the uncertainties with regions. Secondly, our framework considers all unsafe state spaces in the regions and efficiently explore these spaces to reduce risks where the elimination of risks can be formally verified. Thirdly, our framework is well compatible with its learning process, with few adjustments needed for repair. 

We summarize our contributions as follows: (1) We propose a framework for repair of DNNs with respect to input-output safety specifications. Our method does not require safe model references and can successfully repair unsafe DNNs on multiple safety specifications with negligible impact on performance. (2) The method can be utilized with deep reinforcement learning to generate provably safe agents.
(3) We present a novel depth-first-search reachability analysis algorithm that includes both exact and over-approximation methods. This results in a five-fold computational speedup and two-fold memory reduction when compared to other state-of-the-art approaches. 
(4) The framework is evaluated on two benchmark problems where the detailed evolution of model candidates under repair is thoroughly analyzed.  

\section{Preliminaries}
\label{sec:prelim}

\subsection{Reachability Analysis and Set Representation}
Reachability analysis is a process of computing reachable sets for the states of a system w.r.t. an initial state domain. For DNNs, given an input set bounding all possible inputs, reachability analysis computes its output reachable domain. In other words, it computes the domain of all possible outputs that the DNN can produce given an input range. The set normally refers to a convex polytope or a convex region bounded with linear constraints. In this process, sets will be sequentially updated by the affine mapping and activation functions in neurons in the DNNs until the last layer where the final sets compose the reachable domain of the DNN. The choice of set representation is a critical component of reachability analysis algorithms, and it has implications in computational complexity and accuracy of the approach. There are many mathematical structures that enable the definition of a convex polytope. For example, the \textit{half-space representation} defines a polytope as a set of finite linear constraints.  The \textit{vertex representation} defines a polytope with a finite number of extreme points. 
The reachability analysis method in this work mainly relies on two set representations. One is the FVIM~\cite{xiaodong21} for exact reachability analysis, and the other one, a novel over-approximation approach proposed in this work, is the $\mathcal{V}$-zono representation proposed in Section~\ref{sec:reachability-theory}. The new representation improves the computation speed and memory footprint of the algorithm.
In the following, we review the FVIM representation for exact reachability analysis of DNNs~\cite{xiaodong21}.

\subsection{Facet-vertex Incidence Matrix}
A facet-vertex incidence matrix (FVIM) is a complete encoding of the combinatorial structure of a convex set or a polytope~\cite{henk200416}.  It describes the containment relation between facets of a polytope and its vertices, where facets and vertices are types of faces and they are defined in Definition~\ref{def:faces}. The FVIM approach for exact reachability analysis of ReLU DNNs has been shown to be very efficient compared to approaches with different set representations ~\cite{xiaodong21}. One example of the FVIM representation of a 3-dimensional polytope $S$ is shown in Figure~\ref{fig:example-VFVIM}. The polytope contains eight vertices denoted as $\textbf{v}$s and six facets denoted as $F$s. Each facet is a \textit{face} of $S$ which contains four vertices. For instance, the facet $F_1$ denotes the plane containing vertices $\textbf{v}_1$, $\textbf{v}_2$, $\textbf{v}_3$ and $\textbf{v}_4$. The complete containment relation between vertices $\textbf{v}$s and facets $F$s is encoded in the FVIM on the left matrix. Together with real values of vertices, FVIM can represent the set $S$. Formally, it is defined as follows.

\begin{definition}[Faces] Given a polytope $S$ and a supporting hyperplane $\mathcal{H}: a^{\top}x+b=0$ whose halfpsace $a^{\top}x+b\leq0$ or $a^{\top}x+b\geq0$ contains $S$, if the dimension of $F \text{=} \mathcal{H}\cap S$ is $k$, then $F$ is a $k$-dimensional face of $S$ and denoted as $k$-face. A full-dimensional convex polytope $S\subseteq{R}^d$ contain 0-faces,1-faces, \dots, $(d\text{-}1)$-faces which are respectively named vertex, edges, \dots, facets. The cardinality of $k$-faces is denoted as $f_k(S)$.
\label{def:faces}
\end{definition}

\begin{definition}[Facet-vertex incidence matrix] The facet-vertex incidence matrix of a full-dimensional polytope $S\in\mathbb{R}^{d}$ is a matrix $\mathcal{F}\in\{0,1\}^{f_{n\text{-}1}(S)\times f_0(S)}$ where the entry $\mathcal{F}(F,\textbf{v})\text{=}1$ indicates that the facet $F$ contains the vertex $\textbf{v}$, while the entry $\mathcal{F}(F,\textbf{v})\text{=}0$, otherwise. 
\end{definition}

\begin{figure}[H]
    \centering
    \includegraphics[scale=0.7]{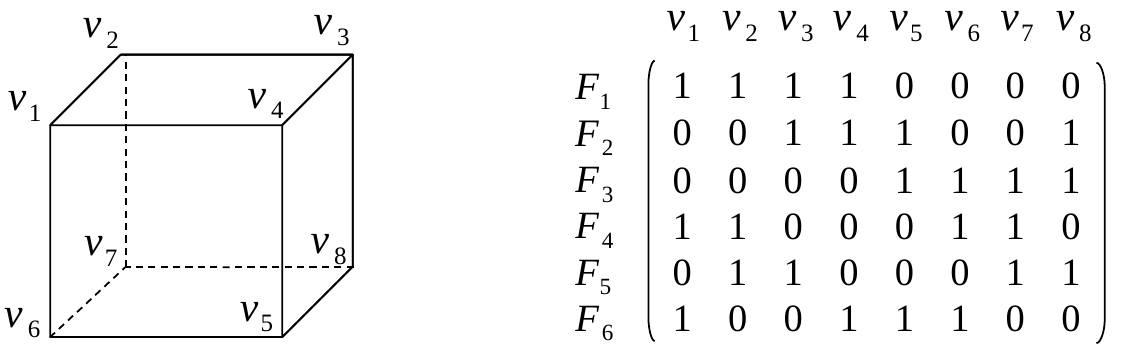}
    \caption{Example of the facet-vertex incidence matrix.}
    \label{fig:FVIM-example}
\end{figure}

Reachability analysis of DNNs with FVIM consists of the sequential application of two main processes. One is affine mapping of the input set by the weights and bias for each layer. This is followed by the transformation operation on the input set through each neuron in the layer. For affine mapping, one useful attribute of FVIM is that it actually only changes the value of vertices and will preserve the FVIM, which can ensure an efficient computation. As the input set passes through neurons, our algorithm checks whether the input range spans over the two linearities of the ReLU function. This is done by computing the lower bound and upper bound of the range. If it spans both linearities, then subsets belonging to different linearities are processed separately. The computation of the lower bound and upper bound of a set is one of primary challenges in the reachability analysis of DNNs. In other works, this problem is commonly encoded with LP solvers~\cite{wong2018provable,zhang2018efficient,tran2019fm,bak2020cav}, which normally deal with a large number of variables and may exhibit undesired efficiency. In contrast, FVIM encodes all the vertices of the set which can be directly used to determine the lower bound and upper bound of the set. Thus the LP problems can be avoided. 


\section{Deep Neural Network Repair}



It has been shown that training of DNNs with adversarial examples is an effective way to improve its robustness with respect to safety~\cite{goodfellow2014explaining, madry2017towards, athalye2018obfuscated, NEURIPS2020_11f38f8e}. These methods utilize a relatively small number of adversarial examples to train more robust DNNs. However, for DNN applications in safety-critical systems, additional guarantees are necessary. It is important to go from robustness improvements to safety guarantees without sacrificing performance of the DNN. 

\subsection{Provably Safe DNNs}
\label{sec:safeDNNs}
 
Let $\mathcal{N}: X \rightarrow Y$ where $X$ and $Y$ are the input and output space be a Deep Neural Network such that given an input $\textbf{x} \in X$, produces an output $\textbf{y}=\mathcal{N}(x)\in Y$. The safety verification problem of DNNs w.r.t. safety properties is formally defined as follows. 

\begin{definition}[Safety Property]
     A safety property $\mathcal{P}$ of a DNN $\mathcal{N}$ specifies an input domain $\mathcal{I} \subseteq X$ and a corresponding unsafe output domain $\mathcal{U} \subseteq Y$. 
     \label{def:safe_prop}
\end{definition}

\begin{definition}[DNN Safety Verification]
      A DNN is safe on a property $\mathcal{P}$, or $\mathcal{N} \models \mathcal{P}$, if for any $\textbf{x}\in\mathcal{I}$ and $\textbf{y}= \mathcal{N}(\textbf{x}) \ \text{then} \ \textbf{y}\notin \mathcal{U}$. Otherwise, it is unsafe, or $\mathcal{N} \not\models \mathcal{P}$. 
     \label{def:Req}
\end{definition}

Given a set of safety properties $\{\mathcal{P}\}^n_{i=1}$, a performance function $\mathcal{A}$, and a candidate DNN $\mathcal{N}$, we define the DNN Repair problem as the problem of retraining or repairing the DNN to generate a new DNN $\mathcal{N'}$ such that all the properties are satisfied and the accuracy or performance of the candidate DNN is maintained. For classification DNNs, the performance function $\mathcal{A}$ refers to the classification accuracy on test data. For DNN agents in DRL, $\mathcal{A}$ refers to the averaged rewards on certain number of episode tests. 

\begin{problem}[DNN Repair]
    Given a DNN candidate $\mathcal{N}$, safety properties $\{\mathcal{P}\}^n_{i=1}$ and performance function $\mathcal{A}$, train a DNN $\mathcal{N}'$ such that $\mathcal{N}' \models \{\mathcal{P}\}^n_{i=1}$ and also $A(\mathcal{N}') - A(\mathcal{N})\geq \varepsilon$. $\varepsilon$ is a constant value used to set the performance threshold. 
    \label{prob:dnn_repair}
\end{problem}





\subsection{Reachability Analysis of DNNs with Backtracking}
\label{sec:reachability}

\begin{figure}[ht]
    \centering
    \includegraphics[scale=0.68]{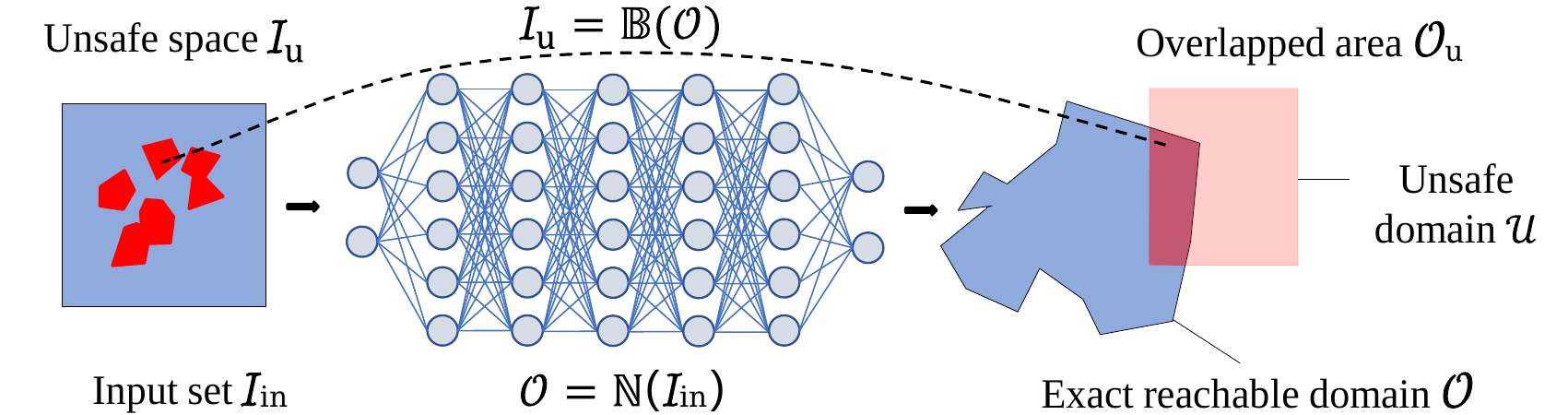}
    \caption{Computation of exact reachable domain and identification of unsafe spaces.}
    \label{fig:reachability}
\end{figure}

At the core of our approach, a reachability analysis method is utilized to determine specification violations. 
While traditional reachability analysis of neural networks focuses on computing output reachable domain given an input domain, 
for neural network repair, it is just as important to backtrack the unsafe reachable domain to the corresponding unsafe input domain containing all adversarial examples.  The input domain that generates unsafe behaviors is then used for the training/repair process of the DNN. The computation of the unsafe input domain is normally associated with the computation of its output reachable domain. The algorithm needs first to determine the overlap between the reachable domain and the predefined unsafe domain before backtracking the corresponding unsafe input space. The computation of output reachable domain as well as the subsequent computation of unsafe input spaces are defined in Definition~\ref{def:reachable_domain} and~\ref{def:unsafe_space}. They are also illustrated in Figure~\ref{fig:reachability}. Given an input set $\mathcal{I}_{in}$, a square input domain in blue, the exact output reachable domain $\mathcal{O}$  can be computed by $\mathcal{O} = \mathbb{N}(\mathcal{I}_{in})$. When $\mathcal{O}$ overlaps with the unsafe domain $\mathcal{U}$ which is $\mathcal{O}_u=\mathcal{O}\cap\mathcal{U}$ and $\mathcal{O}_u\neq \emptyset$, we can compute the unsafe input space $\mathcal{I}_u$ in red area that only contains all the inputs leading to the safety violation.  


\begin{definition}[Output Reachable Domain]
    Let the computation of reachable sets of a DNN $\mathcal{N}$ be denoted as $\mathbb{N}(\cdot)$.
    Given an input set $\mathcal{I}_{in}$ to $\mathcal{N}$, a set of output reachable sets $\{S\}_{k=1}^n$ of $\mathcal{N}$ can be computed as $\{S\}_{k=1}^n = \mathbb{N}(\mathcal{I}_{in})$. We define the output reachable domain as $\mathcal{O}=\bigcup_{k=1}^{n}S_k$.
    \label{def:reachable_domain}
\end{definition}

\begin{definition}[Unsafe Input Domain]
     Given an input set $\mathcal{I}_{in}$ to a DNN $\mathcal{N}$ and its output reachable domain $\mathcal{O}$ that contains reachable sets $\{S\}_{k=1}^n$, if $\mathcal{O}$ overlaps with the unsafe domain $\mathcal{U}$, then it is denoted as $\mathcal{O}_u=\mathcal{O}\cap \mathcal{U}$ and $\mathcal{O}_u$ is named unsafe output reachable domain. The computation of the unsafe input domain refers to the process of computing a set of input subsets $\{I\}_{i=1}^m\subset \mathcal{I}$ and $\mathcal{I}_u = \bigcup_{i=1}^{m}I_i$, such that $\forall \textbf{x} \in \mathcal{I}_u$, its output $\textbf{y}\in \mathcal{O}_u$ and also that $\forall \textbf{x} \notin \mathcal{I}_u$, its output $\textbf{y}\notin \mathcal{O}_u$. This process is denoted as $\mathcal{I}_u = \mathbb{B}(\mathcal{O})$. 
     \label{def:unsafe_space}
\end{definition}

Next, the algorithm for the computation of reachable sets will be presented in detail. 
We assume that DNNs consists of one input layer, multiple hidden layers with ReLU neurons, and one output layer with identity neurons. Except for the input layer, the computation in each layer includes \textit{affine mapping} by the weights and bias ahead and also the process with neurons. \textit{affine mapping} is denoted as $\mathbb{T}(\cdot)$.  The computation function for each neuron is denoted as $\mathbb{E}(\cdot)$. Given an incoming set $S\in \mathbb{R}^d$ to a layer of $l$ neurons, $S$ will be first mapped by $\mathbb{T}(\cdot)$ into $S'\in \mathbb{R}^l$ where the dimension $x_i$ of $\textbf{x} \in S'$ is the input of $i^{th}$ neuron. 

Each ReLU neuron has two different linearities over its input range. For the $i^{th}$ neuron, it can be denotes as $\hat{x}_i = \text{ReLU}(x_i)$ where for $x_i<0$, $\hat{x}_i=0$ and for $x_i\geq 0$, $\hat{x}_i = x_i$. 
For the process $\mathbb{E}_i(S')$ of $S'$ with the $i$th neuron, there are totally three different cases. The first one is that $S'$ only locates in the range $x_i<0$. In this case, the dimension $x_i$ of all $\textbf{x}\in S'$ will be set to 0, which is equivalent to an \textit{affine mapping} on $S'$. The second case is that $S'$ only locates in the range $x_i\geq 0$, where $S'$ will stay unchanged. The third case is that $S'$ spans the two ranges. In this case, $S'$ will be divided into two subsets by a hyperplane $\mathcal{H}: x_i = 0$, with each of subsets lying in one range $x_i<0$ or $x_i\geq 0$. Then, on each subset, either the first or second case is applied.

Overall, the process for one input set with a neuron can generate at most 2 sets. These sets will be subsequently processed with another neuron until all the neurons in the layer are considered. Let the computation of one layer be denoted as $\mathbb{L}(\cdot)$. It then can be formulated as in Equation~\ref{equ:one-layer} where $l$ denotes the number of neurons. The order in which the neurons in a layer are processed is not important. Given an input set $S$, in the worst case, it can output $O(2^l)$ sets. Based on Equation~\ref{equ:one-layer}, the output reachable domain of a DNN can be computed layer by layer as in Equation~\ref{equ:network} where $k$ denotes the number of layers. Suppose the DNN includes $n$ ReLU neurons, it will generate $O(2^n)$ reachable sets. 
\begin{equation}
    \mathbb{L}(S) = (\mathbb{E}_l \circ\dots\circ\mathbb{E}_2 \circ\mathbb{E}_1\circ\mathbb{T})(S)
    \label{equ:one-layer}
\end{equation}
\begin{equation}
    \mathbb{N}(S) = (\mathbb{L}_k\circ\dots\circ\mathbb{L}_2\circ\mathbb{L}_1)(S)
    \label{equ:network}
\end{equation}

Since different linearities of a ReLU neuron are separately considered in each $\mathbb{E}(\cdot)$, the computation of output reachable sets of a DNN is also equivalent to the reachability analysis for linear regions of the DNN. A \textit{linear region} of a piecewise function like ReLU DNNs refers to the maximum convex subset of the input space, on which the function is linear. Taking this fact into account, the work~\cite{xiaodong21} proposes the set representation FVIM to track the connection between reachable sets and their linear regions, such that for any output sets that violates safe properties can be backtracked to its linear region and thus can identify the unsafe input space. 

In order to compute the unsafe input domain, which is needed for DNN repair, we need to first compute the $O(2^n)$ reachable sets. In practice, only a portion of these reachable sets may violate safety specifications and a large amount of the computation is wasted on the safe reachable sets. Therefore, to improve the computational efficiency, we develop a method to filter out such sets and avoid additional computation.

\begin{remark}
    Our reachability analysis algorithm, in the worst case, will require computation of $O(2^n)$ reachable sets with $n$ ReLU neurons in $\mathbb{N}(\cdot)$ and $\mathbb{B}(\cdot)$.
    We aim to develop an over-approximation method to verify the safety of sets computed in Equation~\ref{equ:one-layer}, such that we can filter out the safe set that will not violate the properties $\mathcal{P}$s and avoid unnecessary subsequent computation.
\end{remark}

 To solve this problem, we propose an algorithm that integrates an over-approximation method with the exact analysis method. Over-approximation methods can quickly check the safety of an input set to DNNs. The integration is done as follows. Before an input set $S$ is processed in a layer $\mathbb{L}(\cdot)$, its safety will be first verified with the over-approximation method. If it is safe, it will be discarded, otherwise, it continues with the exact reachability method. Suppose there are $m$ ReLU neurons involved in the computation in Equation~\ref{equ:network}, then it can generate $O(2^m)$ reachable sets whose computation can be avoided if $S$ is verified safe. The integration of the over-approximation algorithm also improves the memory footprint of the algorithm since a large number of sets are discarded early in the process.
 Another problem of the method based on Equation~\ref{equ:one-layer} and \ref{equ:network} is the memory-efficiency issue. 
 As introduced, their computation may take up tremendous amount of the computational memory, may even result in out-of-memory issues, due to the exponential explore of sets. 
 To solve this problem, the algorithm above is designed with the depth-first search, by which the memory usage can be largely reduced. The details of the over-approximation algorithm are presented in Section~\ref{sec:fast_computation}.

\subsection{DNN Repair for Deep Reinforcement Learning}
In deep reinforcement learning (DRL), an agent is replaced with a DNN controller. The inputs to the DNN are states or observations, and their outputs correspond to agent actions. A \textit{property} $\mathcal{P}$ for the DNN agent defines a scenario where it specifies an input state space $\mathcal{I}_{in}$ containing all possible inputs, and also a domain $\mathcal{U}$ of undesired output actions. Here, \textit{safety} is associated with an input-output specification and reachability analysis refers to the process of determining whether a learned DNN agent violates any of its specifications and also the computation of unsafe state domain. This is formally defined as follows.
\begin{definition}[Safe Agent]
     Given multiple safety properties $\{\mathcal{P}_i\}_{i}^n$ for a DNN agent $\mathcal{N}$, the learned agent is safe if and only if for any $\mathcal{P}_i$ , the reachable domain $\mathcal{O}^{[i]}$ for its input state space $\mathcal{I}_{in}^{[i]}$ by $\mathcal{O}^{[i]} = \mathbb{N}(\mathcal{I}_{in}^{[i]})$ does not overlap with its unsafe action space $\mathcal{U}^{[i]}$, namely, $\mathcal{O}^{[i]}\cap\mathcal{U}^{[i]}=\emptyset$.
     \label{def:safe_agent}
\end{definition}

An unsafe agent has the state domains $\mathcal{O}_u$ where their states will result in unsafe actions. Since the traditional adversarial training is usually for regular DNN training algorithms with existing training data, how to utilize such states or adversarial examples in DRL to repair unsafe behaviors remains a problem. 
By considering the fact that DRL learns optimal policies in interactive environments by maximizing the expectation of rewards, one promising way will be introducing penalty to the occurrence of unsafe actions during the learning, such that safety can also be naturally learned from the unsafe state domain. This strategy can also ensure the compatibility with the DRL algorithms, such that they can be seamlessly integrated. Its details are presented in Section~\ref{sec:framework-DRL}.



\section{Framework for DNN Repair}
\label{sec:framework}

\begin{figure}[H]
    \centering
    \includegraphics[scale=0.5]{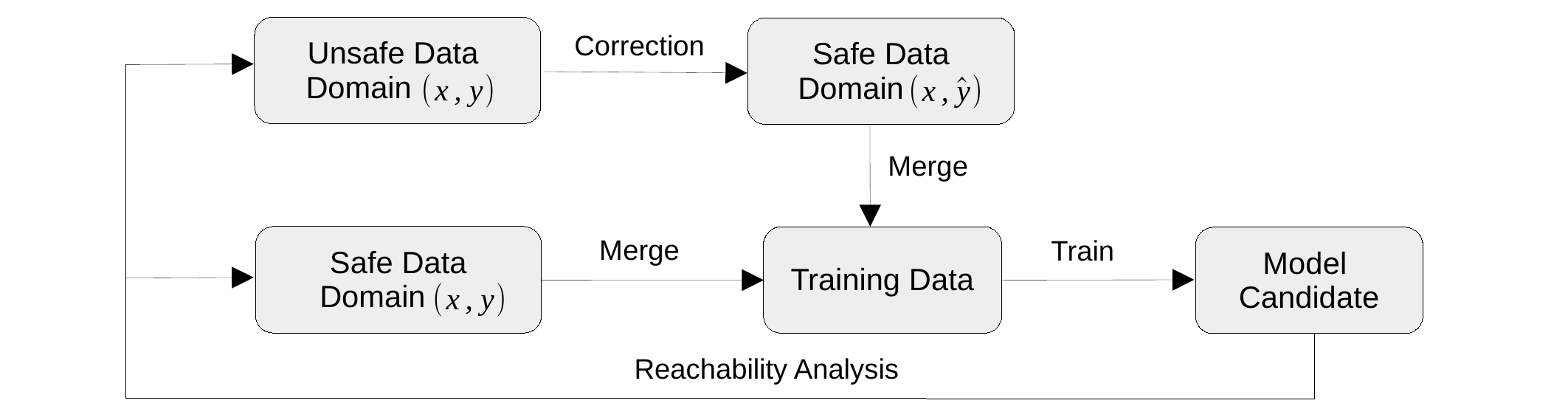}
    \caption{Framework for neural network repair. }
    \label{fig:framework}
\end{figure}

In this section, we propose a solution to Problem \ref{prob:dnn_repair}. The primary idea of our approach is to utilize reachability analysis to incorporate adversarial information into the training process. Different from regular adversarial training which obtains adversarial examples from random attacks, we consider the entire adversarial region by selecting representative examples which are sufficient to represent the region. 
The general approach is shown in Figure~\ref{fig:framework}. 
The retraining process consist of several epochs. For each epoch, reachability analysis, as described in Figure~\ref{fig:reachability}, is conducted to compute the exact unsafe input domain $\mathcal{I}_u$ and its unsafe output reachable domain $\mathcal{O}_u$ for each safety property in $\{\mathcal{P}\}^n_{i=1}$. $\mathcal{I}_u$ and $\mathcal{O}_u$ together are named \textit{unsafe data domain}, which is formally defined in Definition~\ref{def:data_domain}. 
Then data pairs $(\textbf{x}, \textbf{y})$ with $\textbf{x}\in\mathcal{I}_u$ and $\textbf{y}\in \mathcal{O}_u$ which sufficiently represent the unsafe input domain are selected. The selection process is presented next. After determining the representative data pairs, the unsafe output $\textbf{y}$ for a particular adversarial example needs to be corrected before adding it to the original training data. This correction normally requires a safe model as a reference. But in practice, this reference model is usually not available. Therefore, we propose two alternatives for the correction step. One is achieved by editing the unsafe $\textbf{y}$ to its closest safe $\hat{\textbf{y}}$ in the reachable space. The other is for the DRL where unsafe data pairs of the state and action will be penalized through rewards, such that safety can be naturally learned along with the optimal policies. The first case is presented next.

\begin{definition}[Unsafe Data Domain and Unsafe Data Pair]
 Given the unsafe input domain $\mathcal{I}_u$ of a DNN $\mathcal{N}$ on the safety property $\mathcal{P}$ with $\mathcal{I}_u=\bigcup_{k=1}^{m} I_k$, the unsafe reachable domain $\mathcal{O}_u$ is computed by $\mathcal{O}_u=\bigcup_{k}^{m} O_k $ where $O_k = \mathbb{N}(I_k)$ as described in Definition~\ref{def:unsafe_space}. Then the pair $\mathcal{I}_u$ and $\mathcal{O}_u$ is defined as the unsafe data domain of the DNN on the property $\mathcal{P}$, containing unsafe data pairs $(\textbf{x}, \textbf{y})$ with $\textbf{x}\in \mathcal{I}_u$, $\textbf{y}=\mathcal{N}(\textbf{x})\in \mathcal{O}_u$.
 \label{def:data_domain}
\end{definition}

Given a DNN with $n$ safety properties and a set of $l$ training data pair $(\textbf{x}, \textbf{y})$s, the DNN repair problem to satisfy all $n$ properties as well as maintain its performance is formulated as 
\begin{equation}
  \displaystyle{\minimize_{\theta}\Big(\sum_{i=1}^{n} \max_{\textbf{x}\in \mathcal{I}_u^{[i]}} L(f_{\theta}(\textbf{x}), \hat{\textbf{y}}) +  \sum_{j=1}^{l}L(f_{\theta}(\textbf{x}_j), {\textbf{y}_j}) \Big)}
  \label{equ:repair_formula}
\end{equation}
where $f$ denotes DNN, $\theta$ denotes the weight parameters, and $\hat{\textbf{y}}$ represents the optimal safe output for the correction of $\textbf{y}$ in the unsafe data pair $(\textbf{x}, \textbf{y})$ on property $\mathcal{P}_i$, and this correction refers to the \textbf{correction} procedure in Figure~\ref{fig:framework}. Without safe model references for repair, we set $\hat{\textbf{y}}$ to be the closest safe data to $\textbf{y}$ in the space, on which $\|\textbf{y}-\hat{\textbf{y}} \|$ is minimal. 
{The problem of finding $\hat{\textbf{y}}$ can be encoded as a LP problem of finding a $\hat{\textbf{y}}$ on the boundaries of $\mathcal{U}^{[i]}$ such that the distance between $\hat{\textbf{y}}$ and $\textbf{y}$ is minimal,
where the optimal $\hat{\textbf{y}}$ is located on one of its boundaries along its normal vector from $\textbf{y}$}. Let the vector from $\textbf{y}$ to $\hat{\textbf{y}}$ along the normal vector be denoted as $\Delta \textbf{y}$.
Then, the problem of finding $\hat{\textbf{y}}$ can be formulated as 
\begin{equation}
    \hat{\textbf{y}} = \textbf{y} + (1+\alpha)\Delta\textbf{y}, \quad \displaystyle{\min_{\hat{\textbf{y}}\notin \mathcal{U}^{[i]}}\|\textbf{y}-\hat{\textbf{y}} \|}
    \label{equ:correction}
\end{equation}
where $\alpha$ is a very small positive scalar to divert $\hat{\textbf{y}}$ from the boundary of $\mathcal{U}^{[i]}$ into the safe domain.

{The $\textbf{y}\in \mathcal{O}_{u}^{[i]}$ that leads to the maximum loss value for the interior maximization of Equation~\ref{equ:repair_formula} is from the extreme points of $\mathcal{O}^{[i]}_u$, namely, its vertices, because this loss is associated with the maximum distance $\Delta \textbf{y}$ among $\textbf{y}\in\mathcal{O}^{[i]}_u$.}
Let $V_S$ be the set of vertices of $\mathcal{O}^{[i]}_u$, and $V_k$ be the set of vertices of $O_k$ where $O_k = \mathbb{N}(I_k)$. Since $\mathcal{O}^{[i]}_u = \bigcup_{k=1}^{m}O_k$ then $V_S \subseteq \bigcup_{k=1}^{m}V_k$. Recall that an unsafe input set $I_k$ is a \textit{linear region} of the DNN, over which the DNN is linear. Therefore, $O_k$ is essentially an affine mapping from $I_k$ and the vertices of $I_k$ one-to-one correspond to $V_k$ of $O_k$. We can conclude that the vertices of unsafe input sets $\{I\}_{k=1}^m$ contain the optimal $\textbf{x}\in\mathcal{I}_u^{[i]}=\bigcup_{k=1}^{m} I_k$ for the interior maximization of Equation~\ref{equ:repair_formula}. Moreover, vertices are sufficient to represent a convex domain. Therefore, the vertices of unsafe input sets $\{I\}_{k=1}^{m}$ can sufficiently represent the unsafe input domain $\mathcal{I}_u^{[i]}$, and data pairs $(\textbf{x}, \textbf{y})$ where $\textbf{x}$ belongs to the vertices of $\{I\}_{k=1}^{m}$ and $\textbf{y}$ belongs to the vertices $\{O\}_{k=1}^{m}$ can represent the unsafe data domain. These data pairs will be selected to merge into the training data for the adversarial training, which is the \textbf{merge} procedure in Figure~\ref{fig:framework}.

The framework is also described in Algorithm~\ref{al:framework}. To maintain the performance of the repaired DNN, a threshold is also included in Line 5. Function \textbf{reachAnalysis} is used to compute all the safe and unsafe data domains of a DNN on multiple safety properties.  Lines 4 and 7 generate representative data pairs for the adversarial training in Line 10. Function \textbf{Correction} applies a correction on $\textbf{y}$ corresponding to Equation~\ref{equ:correction}.
\begin{algorithm}
    \caption{DNN Repair}
    \label{al:framework}
    \leftline{\textbf{Input}: $\mathcal{N}$, $(\textbf{x}, \textbf{y})_{\textit{training}}$\quad  \comm{\# an unsafe DNN and its training data}}
    \leftline{\textbf{Output}: $\mathcal{N}'$ \quad\comm{\# an safe DNN satisfying all its safety properties with a desired performance}} 
    \begin{algorithmic}[1]
        \Procedure{$\mathcal{N}'$ = Repair}{$\mathcal{N}$}
            \State $\mathcal{N}'\leftarrow \mathcal{N}$
            \While{true}
            \State $\mathcal{D}_{\textit{unsafe}}$, $\mathcal{D}_{\textit{safe}}$ = reachAnalysis($\mathcal{N}$, $\{\mathcal{P}\}_{i=1}^m$) \quad \comm{\# compute unsafe and safe data domains}
            \If{$\mathcal{D}_{\textit{unsafe}}$ is \textit{empty} and $\mathcal{A}(\mathcal{N}')-\mathcal{A}(\mathcal{N})\geq \varepsilon$} \quad \comm{\# $\mathcal{A}$: performance function}
                \State \textbf{break} and \textbf{return} $\mathcal{N}'$
            \EndIf
            \State $(\textbf{x}, \textbf{y})_{\textit{safe}}$, $(\textbf{x}, \textbf{y})_{\textit{unsafe}}$ = Vertices($\mathcal{D}_{\textit{unsafe}}$, $\mathcal{D}_{\textit{safe}}$) \quad \comm{\# representative data pairs}
            \State $(\textbf{x}, \hat{\textbf{y}})$ = Correction($(\textbf{x}, \textbf{y})_{\textit{unsafe}}$)
            \State merge $(\textbf{x}, \hat{\textbf{y}})$,  $(\textbf{x}, \textbf{y})_{\textit{safe}}$ to $(\textbf{x}, \textbf{y})_{\textit{training}}$
            \State $\mathcal{N}'$ = Update($\mathcal{N}'$, $(\textbf{x}, \textbf{y})_{\textit{training}}$)
            \EndWhile
        \EndProcedure
    \end{algorithmic}
\end{algorithm}

\subsection{Framework for Deep Reinforcement Learning}
\label{sec:framework-DRL}
DRL is a machine learning technique where a DNN agent learns in an interactive environment from its own experience. Our framework aims to repair an unsafe agent which violates its safety properties while performance is maintained. The difference of the framework for DRL with the general framework in Figure~\ref{fig:framework} is the correction of $\textbf{y}$ in unsafe data pairs $(\textbf{x}, \textbf{y})$. The correction in this modified framework is achieved by introducing a penalty to the unsafe data pair observed in the learning process, from which safety can be learned. In the following, we introduce the repair framework for DRL.

In a regular learning process, in each \textit{time step}, the agent computes the action and the next state based on the current state. A reward is assigned to the state transition. This transition is denoted as a \textit{tuple} $\langle s,a,r,s'\rangle$ where $s$ is the current state, $a$ is the action, $s'$ is the next state, and $r$ is the reward. Then, this tuple together with previous experience is used to update the agent. The sequence of \textit{time step}s from the beginning with an initial state to the end of the task is called an \textit{episode}. With appropriate parameters settings, the performance of an agent may gradually converge to the optimum after a number of episodes. A good learning also relies on effective policy-learning algorithms. The DRL approach in this work considers one of the most popular algorithms, the deep deterministic policy gradients algorithm (DDPG)~\cite{lillicrap2015continuous} and is utilized on the rocket-lander benchmark~\footnote{https://github.com/arex18/rocket-lander} inspired by the lunar lander~\cite{brockman2016openai}.

\begin{figure}[ht]
    \centering
    \includegraphics[scale=0.45]{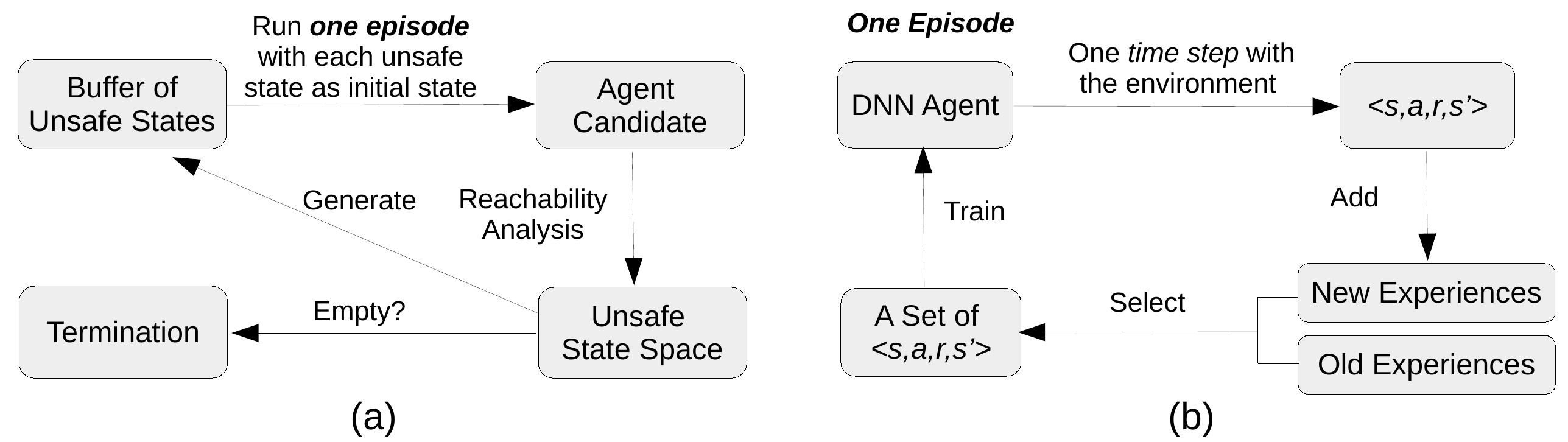}
    \caption{Repair framework for deep reinforcement learning. The loop in (a) represents one epoch. Given an unsafe agent, its unsafe state space is computed with our reachability analysis method, Then, \textit{eposide}s are run with unsafe state as initial states to update the agent, where occurrence of unsafe states will be penalized. In (b),
    the new experiences refer to the experience learned during the repair while the old experiences refer to the ones learned in learning of the original agent. }
    \label{fig:framework-DRL}
\end{figure}

As introduced, the correction of unsafe data pairs $(\textbf{x}, \textbf{y})$ is achieved through self-learning by assigning penalty to unsafe behaviors. Here, \textbf{x} refers to the state input $s$ to the DNN agent and \textbf{y} refers to its output action $a$.  The new framework is shown in Figure~\ref{fig:framework-DRL}. Similar to the general framework, given an unsafe agent candidate in Figure~\ref{fig:framework-DRL}(a), our reachability analysis method computes the unsafe state domain that lead to a wrong action by the agent. The vertices of unsafe state sets $\{I\}_{k=1}^m$ are selected as representative unsafe states for the unsafe domain. The correction of the wrong action $a$ for an unsafe state $s$ will be achieved by running one \textit{episode} with the unsafe state as an initial state as shown in Figure~\ref{fig:framework-DRL}(b). The process of one \textit{episode} is similar to the regular \textit{episode}. The difference is that a new penalty $r$ is incorporated for any unsafe pair $s$ and $a$ in each \textit{time step}. The penalty $r$ is normally being set to the least reward in the old experience, where the \textit{old experiences} refers to the experience from learning the original unsafe agent. In the repair, the $tuple$ in each \textit{time step} will be stored into a global buffer for previous experience, which is named \textit{new experiences}. For training, a set of \textit{tuple}s will be randomly selected from both experiences. The process in Figure~\ref{fig:framework-DRL}(a) will be repeated until the agent becomes safe and its performance is above a predefined threshold. The algorithm is shown in Algorithm~\ref{al:framework-DRL} where Function \textbf{singleEpisode} corresponds to Figure~\ref{al:framework-DRL}(b).

\begin{algorithm}
    \caption{Repair for Deep Reinforcement Learning}
    \label{al:framework-DRL}
    \leftline{\textbf{Input}: $\mathcal{N}$, $E$\quad  \comm{\# an unsafe DNN agent, and its old experience, a set of \textit{tuple}s}}
    \leftline{\textbf{Output}: $\mathcal{N}'$ \quad\comm{\# a safe DNN satisfying all its safety properties with a desired performance}} 
    \begin{algorithmic}[1]
        \Procedure{$\mathcal{N}'$ = Repair}{$\mathcal{N}$}
            \State $\mathcal{N}'\leftarrow \mathcal{N}$
            \While{true}
            \State $\mathcal{D}_{\textit{unsafe}}$ = reachAnalysis($\mathcal{N}$, $\{\mathcal{P}\}_{i=1}^m$) \quad \comm{\# compute unsafe and safe data domains}
            \If{$\mathcal{D}_{\textit{unsafe}}$ is \textit{empty} and $\mathcal{A}(\mathcal{N}')-\mathcal{A}(\mathcal{N})\geq \varepsilon$} \quad \comm{\# $\mathcal{A}$: performance function}
                \State \textbf{break} and \textbf{return} $\mathcal{N}'$
            \EndIf
            \State $S_{\textit{unsafe}}$ = Vertices($\mathcal{D}_{\textit{unsafe}}$) \quad \comm{\# representative unsafe states}
            \For{\textit{s} in $S_{\textit{unsafe}}$}
            \State $\mathcal{N}'$ = singleEpisode($\mathcal{N}'$, \textit{s}, $E$) \quad \comm{\# one episode learning with initial state $s$ and $E$.}
            \EndFor
            \EndWhile
        \EndProcedure
    \end{algorithmic}
\end{algorithm}

\section{Reachability Analysis of DNN}
\label{sec:reachability-theory}
Fast reachability analysis is a core component in our DNN repair framework. 
However, different from traditional algorithms, for DNN repair the emphasis of the algorithm is on finding the unsafe input domain and it's corresponding unsafe output domain. 
Our algorithm builds on the reachability analysis and backtracking method presented in~\cite{xiaodong21}. 
The method utilizes a FVIM set representation for efficient encoding of the combinatorial structure of polytopes. This set representation is suitable for set transformations that are induced by operations in a neural network.
The reachability analysis method presented in~\cite{xiaodong21} is able to compute the output reachable domain of a DNN, and subsequently identify the unsafe input regions. 
However, one disadvantage of the algorithm is that, in the worst case, the number of reachable sets is $O(2^n)$, where $n$ is the number of ReLU neurons.
Its efficiency could be impeded due to this computation of a huge number of sets.


To alleviate this problem, we utilize an novel over-approximation method to speed up computation in Equation~\ref{equ:one-layer} and~\ref{equ:network} by filtering out safe regions in the early stages of the algorithm. 
Since our focus of retraining is on computing the unsafe input domain and it's corresponding unsafe output domain, once we have guarantees of safety for a particular region, we do not need to compute its exact output reachable sets. 
Thereby, the computational efficiency and the memory footprint of the algorithm can be significantly improved.  

Our over-approximation method is based on a new set representation for the linear relaxation of ReLU neurons. 
The new set representation named $\mathcal{V}$-zono is designed to efficiently encode the exponentially increasing vertices of sets in each linear relaxation, and it is totally compatible with the FVIM. 
In the following section, the over approximation with $\mathcal{V}$-zono will be introduced. 
Additionally, to handle the memory-efficiency issue caused by the large amount of sets computed in Equation~\ref{equ:one-layer} and \ref{equ:network}, a depth-first search algorithm is also presented. 

\subsection{Over Approximation with Linear Relaxation}

\begin{figure}[ht]
    \centering
    \includegraphics[scale=0.62]{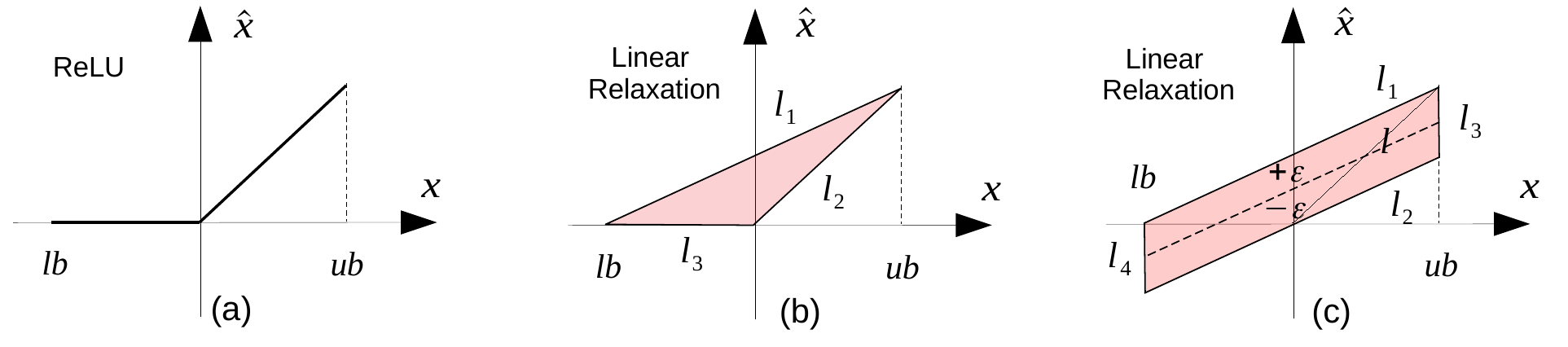}
    \caption{Linear relaxations of ReLU functions with a convex bound: (a) ReLU function, (b) one type of linear relaxation of ReLU function~\cite{wong2018provable}, and (c) linear relaxation utilized in our over-approximation method based on a new set representation. }
    \label{fig:relaxation}
\end{figure}

This section presents our over-approximation method based on the linear relaxation of ReLU. Linear relaxation is commonly used in other related works for fast safety verification of DNNs, such as~\cite{singh2019abstract,gehr2018ai2,zhang2018efficient}. Instead of considering the two different linearities of ReLU neuron over its input in $\mathbb{E}(\cdot)$ of Equation~\ref{equ:one-layer}, these works apply one convex domain to over approximate these linearities to simplify the reachability analysis, as shown in Figure~\ref{fig:relaxation}. This over approximation is named \textit{linear relaxation}. Recall the process of $S'$ with the $i$th neuron in the layer in Section~\ref{sec:reachability}, when the lower bound and the upper bound of the $x_i$ of $\textbf{x}\in S'$ spans the two linearities bounded by $x_i\text{=}0$, $S'$ is supposed to be divided accordingly and their two subsets lying in the range $x_i<0$ or $x_i\geq0$ will be processed in terms of their linearity. The linear relaxation is applied only in such cases as shown in Figure~\ref{fig:relaxation} (b) and (c). When $S'$ only locates in $x_i<0$ or $x_i\geq0$, the computation will be the same as the computation in Section~\ref{sec:reachability}. We denote this process including the linear relaxation as $\mathbb{E}^{app}(\cdot)$, and the computation of one layer as $\mathbb{L}^{app}(\cdot)$. Thus, by simply substituting $\mathbb{E}(\cdot)$ with $\mathbb{E}^{app}(\cdot)$ in Equation~\ref{equ:one-layer}, and substituting $\mathbb{L}(\cdot)$ with $\mathbb{L}^{app}(\cdot)$ in Equation~\ref{equ:network}, we can conduct the over-approximation method shown in Equation~\ref{equ:one-layer-over} and \ref{equ:network-over}. Function $\mathbb{E}^{app}(\cdot)$ only generates one output set for each input set instead of at most two sets. Therefore, given one input set to the DNN, Equation~\ref{equ:network-over} computes one over-approximated output reachable set instead of $O(2^n)$ sets with $n$ ReLU neurons. 

\begin{equation}
    \mathbb{L}^{app}(S) = (\mathbb{E}^{app}_l\circ\dots\circ\mathbb{E}^{app}_2\circ\mathbb{E}^{app}_1\circ\mathbb{T})(S)
    \label{equ:one-layer-over}
\end{equation}
\begin{equation}
    \mathbb{N}(S) = (\mathbb{L}^{app}_k\circ\dots\circ\mathbb{L}^{app}_2\circ\mathbb{L}^{app}_1)(S)
    \label{equ:network-over}
\end{equation}


Here we introduce two of the most common types of linear relaxations for ReLU functions as shown in (b) and (c) of Figure~\ref{fig:relaxation}. The linear relaxation in (b) uses the minimum convex bound. Compared to other linear relaxations, it can over approximate the output reachable domain with the least conservativeness. 
A less conservative relaxation typically leads to more accurate reachability analysis algorithms.
The primary challenge for this relaxation is the estimation of the lower bound $lb$ and the upper bound $ub$ for each ReLU activation function. This is normally formulated as an LP problem. However, since the number of variables equals to the number of activations, solving such problems with traditional methods for each verification may not be tractable. 

One alternative to avoid the LP problems is to use vertices to represent a set, 
This also enables us to easily integrate this representation with the FVIM set representation in the exact reachability analysis. 
One issue with this approach is that doubling of vertices in each ReLU relaxation may add a significant computation cost and memory occupation. The explanation is as follows. Suppose the relaxation is for the $i$th neuron. As shown in (b) and (c), the relaxation introduces an unknown variable $\hat{x}_i$ to the incoming set $S\in \mathbb{R}^d$ and also the relation between $x_i$ and $\hat{x}_i$ bounded in the convex domain. 
\begin{remark}
\label{remark:projection}
The introduction of $\hat{x}_i$ is equivalent to projecting $S$ into  $S_h$ in $(d\text{+}1)$-dimensional space. For each $\textbf{x}\in S$, it will transform into $\textbf{x}_h\in S_h$ with $\textbf{x}_h \text{=} [\textbf{x}; \hat{x}_i]\in \mathbb{R}^{d+1}$. Accordingly, the faces of $S$ will transform into new faces of $S_h$ with increasing their dimension by one. The new faces of $S_h$ are unbounded because of the unknown variable $\hat{x}_i$. The later intersection of $S_h$ with the domain of $x_i$ and $\hat{x}_i$ in the relaxation will yield real values to $\hat{x}_i$.
\end{remark}

For instance, the vertex $\textbf{v}$ of $S$ which is a $0$-dimensional face will turn into $\textbf{v}_h \text{=} [\textbf{v}; \hat{x}_i]$ which is equivalent to an unbounded edge of $S_h$, a $1$-dimensional face. With $S_h$ and the linear relaxation bounds of $x_i$ and $\hat{x}_i$,  $\mathbb{E}^{app}_i(S)$ can be interpreted as the intersection of $S_h$ with these bounds. 

\begin{remark}
\label{remark:intersection}
The convex bounds of the ReLU relaxation consists of multiple linear constraint $l$s. Each $l: \alpha\cdot\textbf{x} + \beta \leq 0$ is one of two halfspaces divided by the hyperplane $\mathcal{H}: \alpha\cdot\textbf{x} + \beta \text{=}0$. The intersection of $S_h$ with each $l$ is essentially identifying the subset of $S_h$ which is generated from division of $S_h$ by $\mathcal{H}$ and locates in the halfspace $l$. 
\end{remark}

Take the (b) relaxation for instance which is bounded by three linear constraints $l_{1}$, $l_{2}$ and $l_{3}$. $\mathbb{E}^{app}_i(S)$ can be formulated as 
\begin{equation}
        \mathbb{E}^{app}_i(S) = S_h\cap\{\textbf{x}\in \mathbb{R}^d\ |\ l_1\cap l_2\cap l_3\}.
    \label{equ:conjunction}
\end{equation}
As introduced above the vertex  $\textbf{v}_h \text{=} [\textbf{v}; \hat{x}_i]$ of $S_h$ is symbolic with $\hat{x}_i$ and is equivalent to an unbounded edge. In a bounded set, an edge includes two vertices. After the intersection of $S_h$ with the linear constraints, $\mathbb{E}^{app}_i(S)$ generates a bounded subset of $S_h$ and meanwhile, each symbolic $\textbf{v}_h$ will yield to two real vertices. Therefore, the vertices of $S_h$ is doubled from the vertices of $S$.

\begin{equation}
    \begin{cases}
       \mathcal{H}_1: (ub-lb)\cdot \hat{x} - ub\cdot x + ub\cdot lb = 0 \\
       \mathcal{H}_2:  (ub-lb)\cdot \hat{x} - ub\cdot x = 0 \\
       \mathcal{H}_3: x - ub = 0 \\
       \mathcal{H}_4:  x - lb = 0 \\
    \end{cases}
    \label{equ:relaxation-li}
\end{equation}

To solve this problem, it is necessary to develop a new set representation that can efficiently encode the exponential explosion of vertices with ReLU relaxations. Here, we choose the relaxation in Figure~\ref{fig:relaxation}(c) because the convex bound for the relaxation is a \textit{zonotope} which can be simply represented by a set of finite vectors and is formulated as a \textit{Minkowski sum}. An efficient representation of vertices can benefit from this simplification. The explanation is as follows. The zonotope in (c) is bounded by two pairs of parallel supporting hyperplanes, $ \mathcal{H}_1, \mathcal{H}_2 $ and $\mathcal{H}_3, \mathcal{H}_4$ as shown in Equation~\ref{equ:relaxation-li}, and their linear constraints are denoted as $l_{1}$, $l_{2}$, $l_{3}$ and $l_{4}$.  Since $l_3$ and $l_4$ are lower and upper bounds of the $x_i$ in $S$ and $S$ itself locates in these constraints, then, the left part of the conjunction in Equation~\ref{equ:conjunction} can be replaced with $\{\textbf{x}\in \mathbb{R}^n\ |\ l_1\cap l_2\}$. For each symbolic vertex $\textbf{v}_h \text{=} [\textbf{v}; \hat{x}_i]$, two new vertices $\textbf{v}_{1}'$ and $\textbf{v}_{2}'$ can be computed from the intersection of the hyperplanes $\mathcal{H}_1$ and $\mathcal{H}_2$ with $\hat{x}_i$ involved. The vertices $\textbf{v}_{1}'$ and $\textbf{v}_{2}'$ are shown in Equation~\ref{equ:new_vertex} where $x_i$ equals to the $v_i$ of $\textbf{v}$.
\begin{equation}
    \textbf{v}_1' = \begin{bmatrix} \textbf{v}\\ \frac{ub\cdot x_i-ub\cdot lb}{ub-lb}\end{bmatrix}, \ \ 
    \textbf{v}_2' = \begin{bmatrix} \textbf{v}\\ \frac{ub\cdot x_i}{ub-lb}\end{bmatrix}
    \label{equ:new_vertex}
\end{equation}

\begin{equation}
    \{\textbf{v}_1', \textbf{v}_2'\} = \Bigg\{\textbf{v}'_c \pm \textbf{v}'_v, \ \ \Bigg| \ \ \textbf{v}'_c = \begin{bmatrix} \textbf{v}\\ \frac{ub\cdot v_i}{ub-lb} - \frac{ub\cdot lb}{2(ub-lb)}\end{bmatrix},\ \ \textbf{v}'_v  = \begin{bmatrix} \textbf{0}\\ \frac{ub\cdot lb}{2(ub-lb)} \end{bmatrix} \Bigg\}
    \label{equ:vv}
\end{equation}
 
These two vertices can also be represented by Equation~\ref{equ:vv} 
where $\textbf{v}'_v$ is a constant vector for the ReLU relaxation $\mathbb{E}_i^{app}(S)$. 
Let the vertices of $S$ be $V$ and $S'\text{=} \mathbb{E}_i^{app}(S)$, then for each $\textbf{v}\in V$ it yields one $\textbf{v}'_c$ for the vertices $V'$ of $S'$.
Let the set of $\textbf{v}_c'$ be denoted as ${V}'_c$, then $V'$ can be represented as Equation~\ref{equ:pre-set-rep} where the doubled vertices are represented by plus-minus with the vector $\textbf{v}'_v$.
\begin{equation}
V' = {V}'_c \pm \textbf{v}'_v
\label{equ:pre-set-rep}
\end{equation}

\begin{figure}
    \centering
    \includegraphics[scale=0.46]{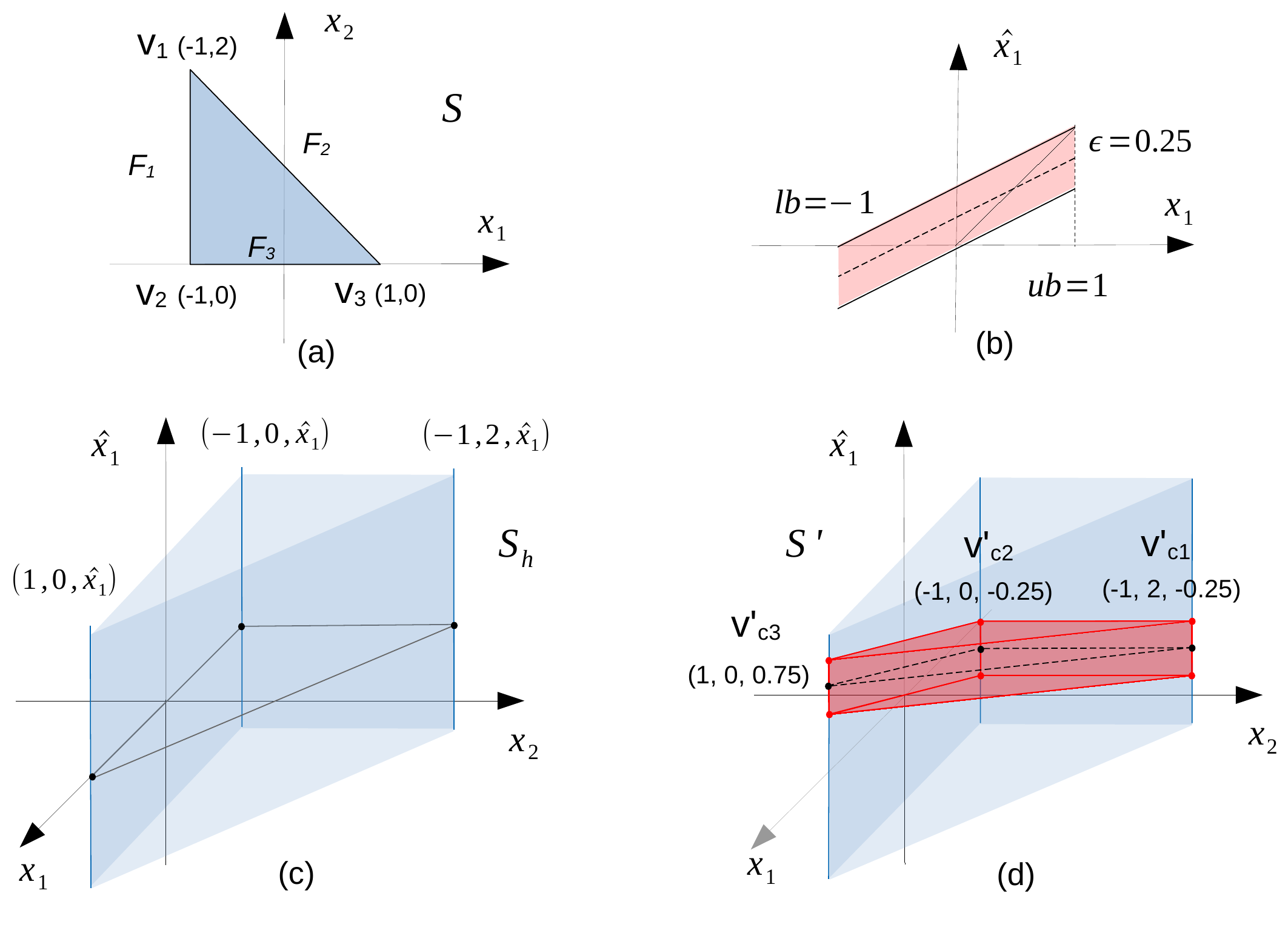}
    \caption{Example of the linear relaxation.}
    \label{fig:example-VFVIM}
\end{figure}

The relaxation is illustrated by in Figure~\ref{fig:example-VFVIM}. The layer includes 2 ReLU neurons. The input set $S$ is 2-dimensional with $\textbf{x}\text{=}[x_1,x_2]^{\top}\in S$, and its vertices $V$ consists of $\textbf{v}_1$, $\textbf{v}_2$ and $\textbf{v}_3$. Here, $x_1$, $x_2$ respectively corresponds to the input of the first neuron and the second neuron. In this example, the process of $S$ w.r.t. the first neuron is demonstrated. We can notice that the lower bound and the upper bound of $x_1$ in $S$ are $lb\text{=}-1$ and $ub\text{=}1$, which indicates that $S$ spans the input range of ReLU function over which the function exhibits two different linearities. Therefore, the linear relaxation is applied in terms of the subfigure (b). As introduced above, there are four linear constraints $l_1$, $l_2$, $l_3$ and $l_4$ bounding this relaxation of $x_1$ and $\hat{x}_1$, whose hyperplanes can be computed by Equation~\ref{equ:relaxation-li}.

The introduction of new variable $\hat{x}_1$ projects 2-dimensional $S$ into 3-dimensional $S_h$ with $\textbf{x}\in S$ transforming into $\textbf{x}_h\text{=}[\textbf{x}; \hat{x}_1]\in S_h$. Accordingly, the vertices $\textbf{v}_1$, $\textbf{v}_2$ and $\textbf{v}_3$ of $S$ transform into new symbolic vertices $[-1,2,\hat{x}_1]^{\top}$, $[-1,0,\hat{x}_1]^{\top}$ and $[1,0,\hat{x}_1]^{\top}$ as shown in the subfigure (c), which are equivalent to three unbounded edges of $S_h$. After the intersection of $S_h$ with those linear constraints, we obtain the final over-approximated set $S'$ represented by the red domain in the subfigure (d) with its 6 vertices represented by the red points. According to Equation~\ref{equ:vv} and \ref{equ:pre-set-rep}, the vertices $V'$ of $S$ can be represented as 
\begin{equation}
    \begin{bmatrix} x_1\\x_2\\\hat{x}_1 \end{bmatrix} \in V', \quad V' = \Bigg\{\textbf{v}'_c\pm \textbf{v}_v'\ \  \Bigg|\ \  \textbf{v}'_c \in \Bigg\{ \begin{bmatrix} -1\\2\\-0.25 \end{bmatrix}, \begin{bmatrix} -1\\0\\-0.25 \end{bmatrix}, \begin{bmatrix} 1\\0\\0.75 \end{bmatrix}\Bigg\}, \ \ \textbf{v}'_v=\begin{bmatrix} 0\\0\\0.25 \end{bmatrix}\Bigg\}
    \label{equ:example-vertices}
\end{equation}
where $\textbf{v}_c'$s are denoted as $\{\textbf{v}'_{c1}, \textbf{v}'_{c2},\textbf{v}'_{c2}\}$ and described by the dark points in (d).

\subsection{Over approximation with \texorpdfstring{$\mathcal{V}-$} -zono}

In the previous section, the a new set representation is preliminarily derived for the linear relaxation in Figure ~\ref{fig:relaxation}(c). This section mainly presents the formal definition of the set representation $\mathcal{V}$-zono, and its utilization in the over approximation of DNNs in Equation~\ref{equ:one-layer-over}. The utilization includes the linear relaxation of ReLU neuron $\mathbb{E}^{app}(\cdot)$, the affine mapping $\mathbb{T}(\cdot)$, as well as the safety verification on the safety properties of DNNs.

The $\mathcal{V}$-zono shares similarities with the $\mathcal{V}$-representation of polytopes introduced in Section~\ref{sec:prelim}, but it can efficiently encode exponentially increasing vertices. It is formally defined in Definition~\ref{def:set-rep}. It consists of \textit{base vertices} $\mathcal{C}$ and \textit{base vectors} $\mathcal{V}$. An example of the vertices representation is demonstrated in Figure~\ref{fig:example-VFVIM}(d). The convex set is the red domain. Its base vertices include three $\textbf{v}'_c$ and the base vectors includes one $\textbf{v}'_v$ as shown in Equation~\ref{equ:example-vertices}. Suppose $\mathcal{C}$ contains $m$ base vertices and $\mathcal{V}$ contains $n$ base vectors, then $\mathcal{C}\pm \mathcal{V}$ efficiently represents $m\times2^n$ vertices in Equation~\ref{equ:set-rep}.

\begin{definition}[$\mathcal{V}$-zono] In the vertices representation, the set $\mathcal{C}$ that contains a set of finite real points $\textbf{v}_c\in \mathbb{R}^d$ is named as Base Vertices, and the set $\mathcal{V}$ that contains a set of finite real vectors $\textbf{v}_v\in \mathbb{R}^d$ is named Base Vectors. Then $\langle \mathcal{C}, \mathcal{V} \rangle$ can represent a convex set $S\subset\mathbb{R}^d$ where $\mathcal{C}\pm \mathcal{V}$ encodes all its vertices. 
\begin{equation}
    \mathcal{C}\pm \mathcal{V} = \{\textbf{v}_c + \sum_{i=1}^{n}(\pm \textbf{v}_{v,i})\ | \  \textbf{v}_c\in \mathcal{C}\ \  \text{and}\ \ \mathcal{V}=\{\textbf{v}_{v,1},\textbf{v}_{v,2},\dots,\textbf{v}_{v,n}\} \}
    \label{equ:set-rep}
\end{equation}
\label{def:set-rep}
\end{definition}

\subsubsection{Linear Relaxation with $\mathcal{V}$-zono}
In the application of $\mathcal{V}$-zono in ReLU relaxation, the vertex computation in $\mathbb{E}_i^{app}(S)$ has been formulated as Equation~\ref{equ:vv}, which deals with regular vertices not represented by $\mathcal{V}$-zono. 
Here, we will formally present the linear relaxation with $\mathcal{V}$-zono where given an input set $S$ in $\mathcal{V}$-zono, the $\mathcal{V}$-zono of $S'=\mathbb{E}_i^{app}(S)$ will be computed. 
Suppose the $\mathcal{V}$-zono of $S$ has base vertices $\mathcal{C}$ and base vectors $ \mathcal{V}$. In terms of Equation~\ref{equ:vv}, a new base vertex $\textbf{v}'_c$ can be computed for each $\textbf{v}$ by 
\begin{equation*}
    \textbf{v}'_c=\begin{bmatrix} \textbf{v}\\ \gamma\end{bmatrix},\ \ \gamma=\frac{ub\cdot v_i}{ub-lb} - \frac{ub\cdot lb}{2(ub-lb)},\ \ \textbf{v}\in V\ \text{and} \ V=\mathcal{C}\pm \mathcal{V}.
\end{equation*}
Each $\textbf{v}'_c$ is computed by incorporating one new dimension to each $\textbf{v}\in V$ with a real value $\gamma$, which is essentially adding one new dimension to each $\textbf{v}_c\in\mathcal{C}$ with $\gamma$, and adding one new dimension to each $\textbf{v}_v\in\mathcal{V}$ with zero. Accordingly, all the new base vertices $\textbf{v}'_c$s can be computed as 
\begin{equation*}
    \Bigg\{ \begin{bmatrix}\textbf{v}_c \\ \gamma \end{bmatrix} + \sum_{i=1}^{n}\Bigg(\pm\begin{bmatrix}\textbf{v}_{v,i} \\ 0 \end{bmatrix}\Bigg)\ \ \Bigg| \  \textbf{v}_c\in\mathcal{C} \ \text{and} \ \textbf{v}_{v,i} \in \mathcal{V}\Bigg\}.
\end{equation*}

Based on the equation above, Equation~\ref{equ:vv} can be extended from the computation of new vertices $\textbf{v}'$s for one $\textbf{v}$ to all $\textbf{v}\in V$. The vertices $V'$ of $S'$ can be computed as below, from which  we can derive the $\mathcal{C}'$ and $\mathcal{V}'$ as shown in Equation~\ref{equ:new-rep}.
\begin{equation*}
     \Bigg\{\begin{bmatrix}\textbf{v}_c \\ \gamma \end{bmatrix} + \Bigg(\sum_{i=1}^{n}\pm\begin{bmatrix}\textbf{v}_{v,i} \\ 0 \end{bmatrix}\Bigg) \pm \textbf{v}'_v \ \ \Bigg| \ \textbf{v}_c\in\mathcal{C}, \ \text{and}\ \textbf{v}_{v,i} \in \mathcal{V}\Bigg\}
\end{equation*}
\begin{equation}
    \mathcal{C}' = \Bigg\{\begin{bmatrix}\textbf{v}_c \\ \gamma \end{bmatrix}\ \Bigg|\ \textbf{v}_c \in \mathcal{C}\Bigg\}, \quad
    \mathcal{V}' =  \Bigg\{\begin{bmatrix}\textbf{v}_{v} \\ 0 \end{bmatrix},\ \ \textbf{v}'_v \ \ \Bigg| \ \ \textbf{v}_{v} \in \mathcal{V}\Bigg\}
    \label{equ:new-rep}
\end{equation}

From Equation~\ref{equ:new-rep}, we notice that with the linear relaxation of each ReLU neuron $\mathbb{E}^{app}_i(\cdot)$, the dimension of vertices in $\mathcal{V}$-zono will be increased by one because by introducing the new dimension or variable $\hat{x}_i$ the old dimension $x_i$ still remains. It will result in the dimension inconsistency with the subsequent affine mapping between layers. This old dimension can be eliminated with projecting the set $S'$ on it by replacing the old dimension $x_i$ in the vertices with the new $\hat{x}_i$. The projection is reflected on Equation~\ref{equ:new-rep2} with the updated $\mathcal{C}'$ and $\mathcal{V}'$.
\begin{equation}
    \mathcal{C}' = \{\textbf{v}_c\ |\ \forall \textbf{v}_c \in \mathcal{C},\ \textbf{v}_c[i]=\gamma \}, \quad
    \mathcal{V}' =  \{\textbf{v}_{v} ,\ \textbf{v}'_v \ | \ \forall \textbf{v}_{v} \in \mathcal{V},\ \textbf{v}_v[i]=0\}
    \label{equ:new-rep2}
\end{equation}

After the projection, part of the points $\mathcal{C}'\pm\mathcal{V}'$ will become the actual vertices of the projected set and the rest will become its interior points. 
The projection of a polytope $S$ into a lower-dimensional space will generate another polytope $S_l$ whose every face is a projection of a face of $S$. It indicates that the vertices $V_l$ of $S_l$ are from the projection of subset of the vertices $V$ of $S$. Since after the projection Equation~\ref{equ:new-rep2} preserves all projected vertices from Equation~\ref{equ:new-rep}, there are redundant points in Equation~\ref{equ:new-rep2} which are not vertices but only interior points of the projected polytope. This redundancy is allowed in the vertices representations of polytopes, as well as our set representation $\mathcal{V}$-zono. 
The detection and elimination of this redundancy requires and additional algorithm, which is out of the scope of this work and will be our future work.
In addition, since $\mathcal{V}$-zono can efficiently encode vertices, the redundancy issue will not greatly affect the efficiency of the algorithm, which is also demonstrated in the experiments.

\begin{figure}[H]
    \centering
    \includegraphics[scale=0.46]{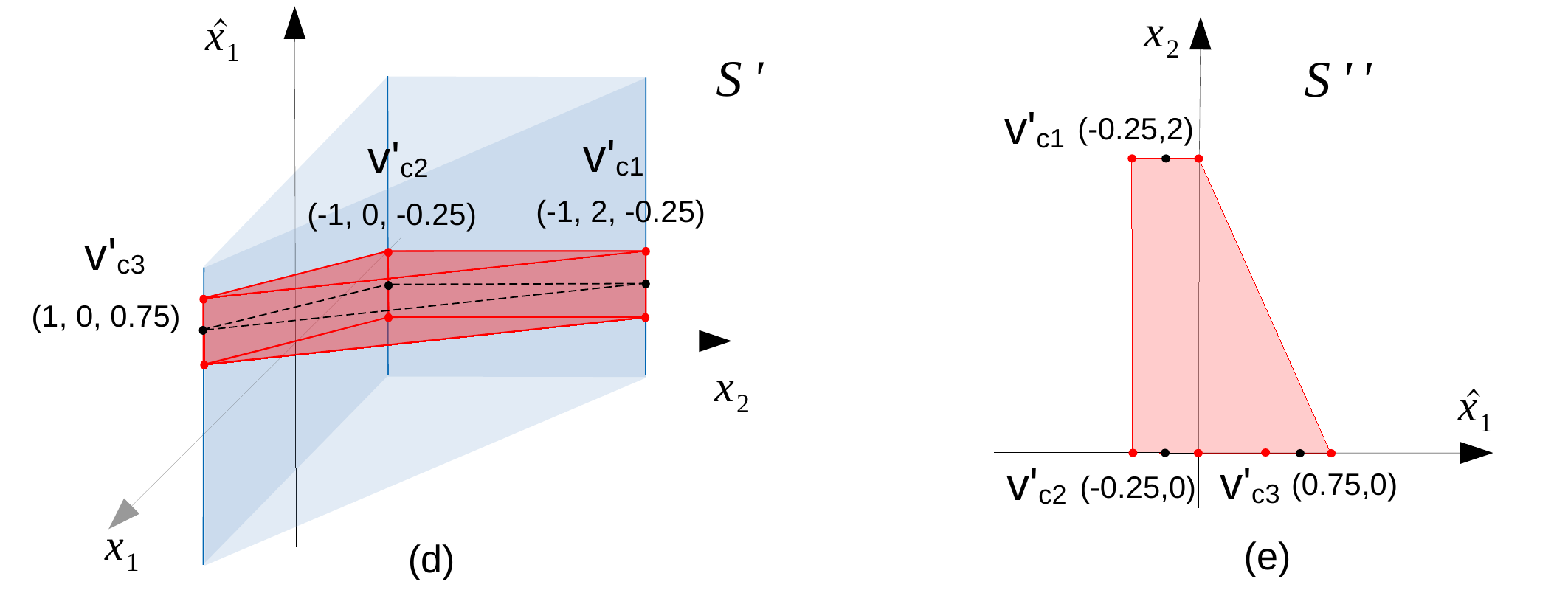}
    \caption{Example of the linear relaxation.}
    \label{fig:Vzono-projection}
\end{figure}

An example of the projection of the set $S'$ in Figure~\ref{fig:relaxation}(d) is shown in Figure~\ref{fig:Vzono-projection}. In the ReLU relaxation, the $x_1$ is the old dimension and $\hat{x}_1$ is the new one, therefore, the set will be projected on the dimension $x_1$ which will maintain the exact bounded relation between $\hat{x}_1$ and $x_2$. The projection generates another polytope represented by the red domain in (e). Its $\mathcal{V}$-zono for the vertices is 
\begin{equation}
     \begin{bmatrix} \hat{x}_1\\x_2 \end{bmatrix}\in V'',\quad V'' = \bigg\{\textbf{v}''_c\pm \textbf{v}_v''\ \  \bigg|\ \  \textbf{v}''_c \in \bigg\{ \begin{bmatrix} -0.25 \\2 \end{bmatrix}, \begin{bmatrix} -0.25\\0 \end{bmatrix}, \begin{bmatrix} 0.75 \\0 \end{bmatrix}\bigg\}, \ \ \textbf{v}''_v=\begin{bmatrix} 0.25\\0 \end{bmatrix}\bigg\}.
    \label{equ:example-vertices2}
\end{equation}
As shown in (d), its $\mathcal{V}$-zono contains 6 vertices denoted as red dots. After the projection as shown in (e), the new $\mathcal{V}$-zono contains 6 points which are the projections of these 6 vertices. 4 of them are actual the vertices of $S''$ and the rest 2 points are redundant.

\subsubsection{Affine Mapping with $\mathcal{V}$-zono}
As shown in Equation~\ref{equ:one-layer-over}, the over approximation of reachable domain also includes affine mapping $\mathbb{T}_{(W,b)}(\cdot)$ between layers by weights $W$ and bias $b$. As introduced in Section~\ref{sec:prelim}, affine mapping on a set $S$ only changes the value of its vertices. Suppose the vertices $V$ of $S$ is represented by $\mathcal{C}\pm \mathcal{V}$, then $\mathbb{T}_{(W,b)}(S)$ on $V$ can be formulated as 
\begin{align}
\begin{split}
    W\cdot V + b & = W\cdot(\mathcal{C}\pm \mathcal{V}) + b \\
    &= W\cdot \mathcal{C} + b \pm W\cdot\mathcal{V}.
\end{split}
\end{align}
The new $\mathcal{C}'$, $ \mathcal{V}'$ for new vertices $V'$ after the affine mapping are
\begin{align}
\begin{split}
    \mathcal{C}' = W\cdot \mathcal{C} + b & = \{W\cdot\textbf{v}_c +b \ |\  \textbf{v}_c\in \mathcal{C}\}\\
    \mathcal{V}' = W\cdot \mathcal{V} & = \{W\cdot\textbf{v}_v\ |\  \textbf{v}_v\in \mathcal{C}\}.
\end{split}
\label{equ:Vzono-affine}
\end{align}

\subsubsection{Safety Verification with $\mathcal{V}$-zono}
The safety verification problem is to determine whether an output reachable domain overlaps with the unsafe domain bounded by linear constraints. Let one linear constraint be denoted as $l: \alpha^{\top}\cdot \textbf{x}+\beta \leq 0$. Suppose vertices $V$ of the over approximated output domain are represented by $\mathcal{C}\pm\mathcal{V}$ and $\mathcal{V}=\{\textbf{v}_{v,1}, \textbf{v}_{v,2},\dots,\textbf{v}_{v,n}\}$. Then, the verification of $V$ w.r.t. the linear constraint $l$ can transform into checking if the minimum value of $\alpha^{\top} \textbf{v}+\beta$ over the vertices $\textbf{v}\in V$ is not positive, which is formulated as 
\begin{equation}
    \minimize (\alpha^{\top}(\mathcal{C}\pm \mathcal{V}) + \beta).
    \label{equ:minimize}
\end{equation}
The internal formula can be extended as 
\begin{align*}
    \begin{split}
        \alpha^{\top}(\mathcal{C}\pm \mathcal{V}) + \beta & = \alpha^{\top}\mathcal{C} + \beta \pm \alpha^{\top}\mathcal{V}
        = \alpha^{\top}\mathcal{C} + \beta + \sum_{i=1}^{k}\pm (\alpha^{\top}\textbf{v}_{v}).
    \end{split}
    \label{equ:}
\end{align*}
Then for each $\textbf{v}_c\in\mathcal{C}$, we can compute its minimum $\alpha^{\top}\textbf{v}_c+\beta +\sum_{i=1}^{k}\text{-}|\alpha^{\top}\textbf{v}_{v}|$. By computing the minimums of all $\textbf{v}_c\in\mathcal{C}$, we can determine the global minimum value in Equation~\ref{equ:minimize} and thus complete the safety verification.

\subsection{Fast Computation of Unsafe Input Spaces of DNNs}
\label{sec:fast_computation}
In the previous sections, an over-approximation method for the reachability analysis of DNNs based on the linear relaxation of ReLU neurons is developed. The method is formulated as Equation~\ref{equ:one-layer-over} and \ref{equ:network-over}.  As introduced, it is utilized to integrate with the exact analysis method formulated in Equation~\ref{equ:one-layer} and \ref{equ:network} to filter out all the safe intermediate sets that are computed in each $\mathbb{L}(\cdot)$ and are not necessary for the computation of unsafe input spaces for DNNs.  In this section, the algorithm for such integration will be presented. This algorithm is based on the depth-first search to handle the memory-efficiency issue due to large amount of sets computed in each layer. 

Algorithm~\ref{al:all} describes the integration of the computation of unsafe input spaces of DNNs with the proposed over-approximation method. Given an input domain $S$, it can compute all the unsafe input spaces w.r.t. the safety properties of the DNN. The details of each function are as follows.
\begin{enumerate}
    \item Function $\textbf{Reach}(\cdot)$ is a recursive function which, in each recursion, computes only one of input sets generated from the last layer to reduce the burden on the computational memory. Its base case is Line 3-5 where the computation reaches to the last layer of the DNN and unsafe input spaces will be computed. The recursion depth is the number of the DNN layers
    \item Function $\textbf{outputOverApp}(\cdot)$ over approximates the output domain of the DNN for each input set which is computed from the last layer with the exact analysis method in Equation~\ref{equ:one-layer} and \ref{equ:network}. This function corresponds to Equation~\ref{equ:one-layer-over} and \ref{equ:network-over}. The details is shown in Algorithm~\ref{al:overapp}. In the beginning, the set $S$ represented by FVIM is transformed into the $\mathcal{V}$-zono in Line 2. Subsequently, in each layer, it will be first processed by the affine mapping in Line 4 which corresponds to the $\mathbb{T}(\cdot)$ in Equation~\ref{equ:one-layer-over} and the computation of $\mathcal{C}'$ and $\mathcal{V}'$ in Equation~\ref{equ:Vzono-affine}. Then, in Line 5, it will be processed with ReLU neurons in the layer based on the linear relaxation. This process corresponds to the each $\mathbb{E}^{app}$ in Equation~\ref{equ:one-layer-over} and also the computation of $\mathcal{C}'$ and $\mathcal{V}'$ in Equation~\ref{equ:new-rep2}.
    \item Function $\textbf{safetyCheck}(\cdot)$ checks the safety of the over-approximated output domain computed in Line 6 with respect to safety properties. This function corresponds to Equation~\ref{equ:minimize} where the vertices of the output domain are checked with each linear constraints of the unsafe domain defined in the properties. It returns \textit{unsafe} if any vertex satisfies in all the constraints, otherwise, \textit{safe}. When $S$ is verified safe, the computation in the following layers can be abandoned because $S$ will not lead to safety violation. 
    \item Function $\textbf{layerOutput}(\cdot)$ computes the reachable sets for the current layer with input sets computed from the previous layer, which corresponds to Equation~\ref{equ:one-layer} and \ref{equ:network}. All sets in this computation are represented by FVIMs. The details is shown in Algorithm~\ref{al:one-layer}.
\end{enumerate}

The computational complexity of Algorithm~\ref{al:all} is that given an input set to a DNN with $n$ ReLU neurons, $O(2^n)$ output reachable sets will be computed. In practice, the utilization of the over approximation method can significantly reduce the amount and improve the computational efficiency. The experimental results indicate that Algorithm~\ref{al:all} can be around five times faster than the algorithm without the over-approximation method. 
\begin{algorithm}
    \caption{Computation of unsafe input spaces of a neural network}
    \label{al:all}
    \leftline{\textbf{Input}: $S$ \quad  \comm{\# one input set to the neural network}}
    \leftline{\textbf{Output}: $\mathcal{O}_{unsafe}$ \quad\comm{\# $\mathcal{O}_{unsafe}$: unsafe input spaces of the DNN}}
    \begin{algorithmic}[1]
        \Procedure{$\mathcal{O}_{unsafe}$ = Reach}{\textit{S}, \textit{layer}} \quad \comm{\# $S$: an input set; $layer$: the layer ID }
            \State $\mathcal{O}_{unsafe} = empty$ 
            \If{\textit{layer} == \textit{lastlayer}} \quad\comm{\# \textit{lastlayer}: the ID of the last layer}
                \State \textit{unsafety} = Backtrack(\textit{S}) \quad\comm{\# \textit{unsafety}: unsafe input space for the unsafe domain in $S$}
                \State \textbf{return} \textit{unsafety}
            \EndIf
            \State \textit{overapp} = outputOverApp(\textit{S}) \quad\comm{\# \textit{overapp}: over approximated output domain of the DNN}
            \If{safetyCheck(\textit{overapp})}
                \State \textbf{return} None
            \EndIf
            \State $\mathcal{O}_c$ = layerOutput(\textit{S}, \textit{layer}) \quad\comm{\# $\mathcal{O}_{c}$: output reachable sets of the current layer for \textit{S}}
            \For{\textit{S} in $\mathcal{O}_{c}$}
                \State $\mathcal{O}_{unsafe}$.extend(Reach(\textit{S}, \textit{layer}+1))
            \EndFor
            \State \textbf{return} $\mathcal{O}_{unsafe}$
        \EndProcedure 
    \end{algorithmic}
\end{algorithm}

\begin{algorithm}
    \caption{Reachable-domain Over approximation of a neural network}
    \label{al:overapp}
    \leftline{\textbf{Input}: $S$ \quad  \comm{\# one input set to the current layer}}
    \leftline{\textbf{Output}: $\mathcal{O}$ \quad\comm{\# one over approximated output reachable set of the current layer}} 
    \begin{algorithmic}[1]
        \Procedure{$\mathcal{O}$ = outputOverApp}{\textit{S}, \textit{layer}}
        \State \textit{S} = Vzono(\textit{S}) \quad\comm{\# transform the FVIM representation of $S$ to the $\mathcal{V}$-zono}
        \For{\textit{layer} = 1:\textit{lastlayer}}
            \State \textit{S} = affineMapping(\textit{S}, \textit{layer}) \quad\comm{\# update base vertices and base vectors}
            \State \textit{S} = reluLayerRelaxation(\textit{S}) \quad\comm{\# relaxation of ReLU neurons}
        \EndFor
        \State \textbf{return} $\mathcal{O}\leftarrow S$
        \EndProcedure
    \end{algorithmic}
\end{algorithm}

\begin{algorithm}
    \caption{Reachable set computation of one layer}
    \label{al:one-layer}
    \leftline{\textbf{Input}: $S$ \quad  \comm{\# one input set to the current layer}}
    \leftline{\textbf{Output}: $\mathcal{O}$ \quad\comm{\# output reachable sets of the current layer}} 
    \begin{algorithmic}[1]
        \Procedure{$\mathcal{O}$ = layerOutput}{\textit{S}, \textit{layer}}
            \State $\mathcal{O}==empty$
            \State \textit{S} = affineMapping(\textit{S}, \textit{layer})
            \If{\textit{layer} == \textit{lastlayer}}
                \State\textbf{return} \textit{S}
            \EndIf
            \State $\mathcal{O}$.extend(reluLayer(\textit{S})) \quad\comm{\# compute exact reachable sets with each ReLU neuron}
            \State \textbf{return} $\mathcal{O}$
        \EndProcedure
    \end{algorithmic}
\end{algorithm}


\section{Experiments and Evaluation}
In this section, we evaluates the performance of the framework, including the performance of the reachability analysis method.
Recall from Section~\ref{sec:framework} that we consider two alternatives for the correction of unsafe data in the absence of a safe model reference. 
The first case considers transformations to the unsafe data points to the nearest safe set. The second case incorporates repair as part of the learning process. 
We evaluate the first case with a well-known benchmark named HorizontalCAS. HorizontalCAS is an airborne collision avoidance system, part of the ACAS Xu family proposed by~\cite{julian2019guaranteeing}. HorizontalCAS has neural networks as controllers. The code and training data for the benchmark are publicly available~\footnote{https://github.com/sisl/HorizontalCAS}, based on which we train all DNNs. Unsafe DNNs will be first identified from these DNNs for the further repair with our framework.
For the second approach, we apply our framework for repair in safe deep reinforcement learning on a well-known benchmark: the rocket lander based on the lunar lander~\cite{brockman2016openai}. The hardware configuration is Intel Core i9-10900K CPU @3.7GHz$\times$, 10-core and 20-thread Processor, 128GB Memory, 64-bit Ubuntu 18.04.

In addition to comparing with the related work~\cite{xiaodong21}, our experimental evaluation examines whether the repairing process can converge on the multiple safety properties, how repairing unsafe behaviors on one property will affect other properties, and how the correction of unsafe data to its closest safe data affects the behavior of repaired DNNs. 

\subsection{HorizontalCAS DNN Controller Repair}
The original controller for the HorizontalCAS is based on a Markov Decision Process (MDP) with large numeric tables~\cite{julian2019guaranteeing}. These controllers are replaced with neural networks. This reduces the memory footprint significantly. There are five continuous inputs and two discrete inputs, For each combination of the two discrete input values, one neural network is trained. Overall, the controller consists of an array of 45 feed-forward neural networks. Each neural network is denoted as $N_{ij}$ where $i$ is an integer index ranging in $[1,5]$ and $j$ is an integer index ranging in $[1,9]$. The input to each DNN is a 5 dimensional continuous sensor measurement of the dynamics between the ownship and the intruder. The inputs are denoted as $[\rho(\text{feet}), \theta(\text{deg}), \psi(\text{deg}), v_{\text{own}}(\text{feet/s}),  v_{\text{int}}(\text{feet/s})]$ and they are, respectively, the distance between ownship and intruder, the angle of the ownship heading direction relative to intruder, angle of intruder heading direction relative to ownship heading direction, velocity of ownship and velocity of intruder. The lower bound and upper bound of their ranges is as follows:
\begin{equation*}
    lb = [0, -\pi, -\pi, 100, 0]; \quad ub = [56000, \pi, \pi, 1000, 1000].
\end{equation*}
There are 5 outputs corresponding to 5 action advisories which are, respectively, clear of conflict, weak right, strong right, weak left and strong left. The action with the maximum output will be selected. Each neural network includes 300 ReLU neurons which are fully connected. And for each neural network, there are several safety properties defined. In each safety property, for an input domain, desired action advisories are defined to avoid aircraft collision. 

All 45 neural networks are well trained with the provided training data and default parameter settings in their code. The accuracy of the DNNs is over 94\%. Here, we design three safety properties for all the networks in terms of the safety properties in work~\cite{katz2017reluplex}. They are as follows:
\begin{enumerate}
    \item \textit{Property 1}: for the input constraints $\rho\geq 50000$, $v_{\text{own}}\geq 900$ and $v_{\text{int}}\leq 60$, the desired output should be located in the domain where the output of the action advisory \textit{clear-of-conflict} should not be the minimum. Then the unsafe output domain will be $y_1\leq y_2\cap y_1\leq y_3\cap y_1\leq y_4\cap y_1\leq y_5$.
    \item \textit{Property 2}: for the input constraints $1500\leq\rho\leq1800$, $-0.06\leq\theta\leq0.06$, $\psi\geq3.10$, $v_{\text{own}}\geq880$ and $v_{\text{int}}\geq860$. The desired output should that the action advisory \textit{clear-of-conflict} should not be the minimum.
    \item \textit{Property 3}: for the input constraints $1500\leq\rho\leq1800$, $-0.06\leq\theta\leq0.06$, $\psi=0$, $v_{\text{own}}\geq900$ and $v_{\text{int}}\geq700$. Their desired output is should that the action advisory \textit{clear-of-conflict} should not be the minimum.
\end{enumerate}
As introduced in Section~\ref{sec:framework}, the correction of unsafe data pairs $(\textbf{x}, \textbf{y})$ is achieved by changing unsafe $\textbf{y}$ to its closest safe $\hat{\textbf{y}}$ by adding a vector $\Delta\textbf{y}$, which is $\hat{\textbf{y}} = \textbf{y} + \Delta\textbf{y}$. The vector $\Delta\textbf{y}$ represents the normal vector with the minimum length, among the normal vectors $\alpha^{\top}$s of the boundary hyperplanes $\alpha^{\top} \textbf{x} + \beta=0$ of the unsafe domain. In the safety properties, we have a common unsafe domain $\mathcal{U}$ bounded by 4 linear constraints and their boundary hyperplanes are respectively, $y_1-y_2=0$, $y_1-y_3=0$, $y_1-y_4=0$ and $y_1-y_5=0$. For a $\textbf{y}\in \mathcal{U}$, its $\Delta\textbf{y}$ is computed over these hyperplanes. 

The safety of neural networks is first verified with our reachability analysis method. They are determined \textit{safe} if there is no unsafe input space computed on all three safety properties. Then, for unsafe networks that violate at least one of the properties, we conduct the repair process with our framework. The repair process monitors all the properties in case that the model candidate turns unsafe on new properties. 
11 of the 45 networks are verified unsafe. The parameter setting for the training of the model in the framework is the same as the ones for the training of the original DNNs. The performance threshold is set to 93\%. The experimental results are shown in Table~\ref{tab:HorizontalCAS}. We can see that all the unsafe networks are successfully repaired by our framework, and that compared to the original model, the accuracy changes on the repaired safe model are negligible. There is no obvious performance degradation on the repaired models. This may be because the $\Delta \textbf{y}$s for the correction is trivial in these cases, seldom impacting accuracy. We can also notice that the network is repaired efficiently in 3 epochs within totally one minute in most of the cases. It is noteworthy that each repair of the violation on one property does not induce violations on other properties. This is likely due to the fact that the input regions defined in the safety properties are apart from each other and repair of unsafe behaviors over one region hardly affects the correct behaviors over other regions.

\begin{table}[h]
\caption{Repair of neural network controllers for HorizontalCAS. There are 11 unsafe neural networks.  \textbf{accuracy changes} represents the accuracy difference (\%) between the repaired safe model and the original unsafe model. \textbf{epochs of repair} represents the number of repair iterations. \textbf{running time} represents the computational time for the repair. }
\resizebox{13cm}{!}{
\begin{tabular}{lccccccccccc} \toprule
Neural Networks      & $N_{11}$ & $N_{12}$ & $N_{15}$ & $N_{16}$ & $N_{17}$ & $N_{19}$ & $N_{26}$ & $N_{41}$ & $N_{52}$ & $N_{55}$ & $N_{59}$ \\ \midrule
\textbf{Accuracy Changes(\%)} & +0.55     & +0.75     & +1.3      & +0.39     & +1.54     & +0.85     & +2.25     & +0.45     & -0.31    & +0.74     & +0.39     \\
\textbf{Epochs of repair}  & 3        & 3        & 2        & 3        & 14        & 3        & 11       & 2        & 2        & 2        & 2        \\
\textbf{Runing Time(sec)}     & 24.3     & 23.1     & 15.2     & 31.5     & 1086.4    & 59.3     & 504.2    & 10.2     & 11.1     & 20.7     & 7.9     \\ \bottomrule
\end{tabular}}
\label{tab:HorizontalCAS}
\end{table}

\begin{figure}[ht]
    \centering
    \begin{subfigure}[t]{0.22\textwidth}
        \centering
        \includegraphics[scale=0.28]{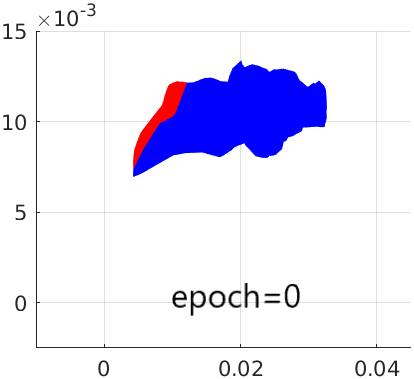}
    \end{subfigure}
    \begin{subfigure}[t]{0.22\textwidth}
        \centering
        \includegraphics[scale=0.28]{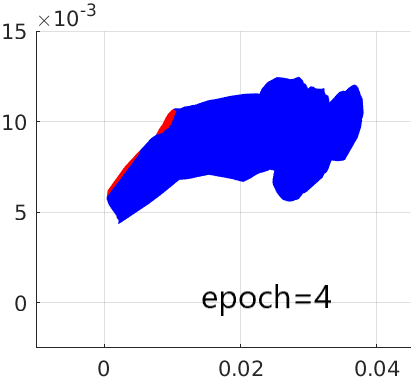}
    \end{subfigure}
    \begin{subfigure}[t]{0.22\textwidth}
        \centering
        \includegraphics[scale=0.28]{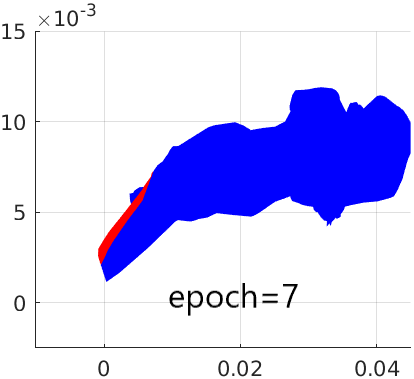}
    \end{subfigure}
    ~
    \begin{subfigure}[t]{0.22\textwidth}
        \centering
        \includegraphics[scale=0.28]{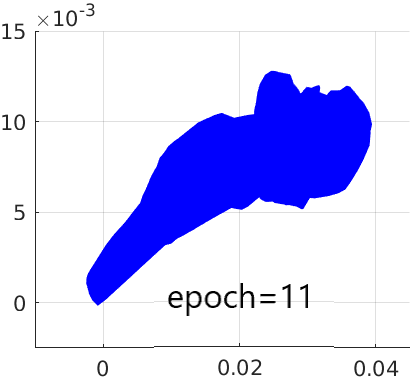}
    \end{subfigure}
    \caption{The evolution of the output reachable domain and the unsafe reachable domain in the repair of DNN $N_{26}$ on property 1. The domains are projected on the output $y_1$ and $y_2$ which are respectively, the $x$ axis and $y$ axis. The blue area represents the exact output reachable domain while the red area represents the unsafe reachable domain which is a subset of the exact output reachable domain.}
    \label{fig:HorizontalCAS_sets}
\end{figure}

\begin{figure}[ht]
    \centering
    \begin{subfigure}[t]{0.42\textwidth}
        \centering
        \includegraphics[scale=0.5]{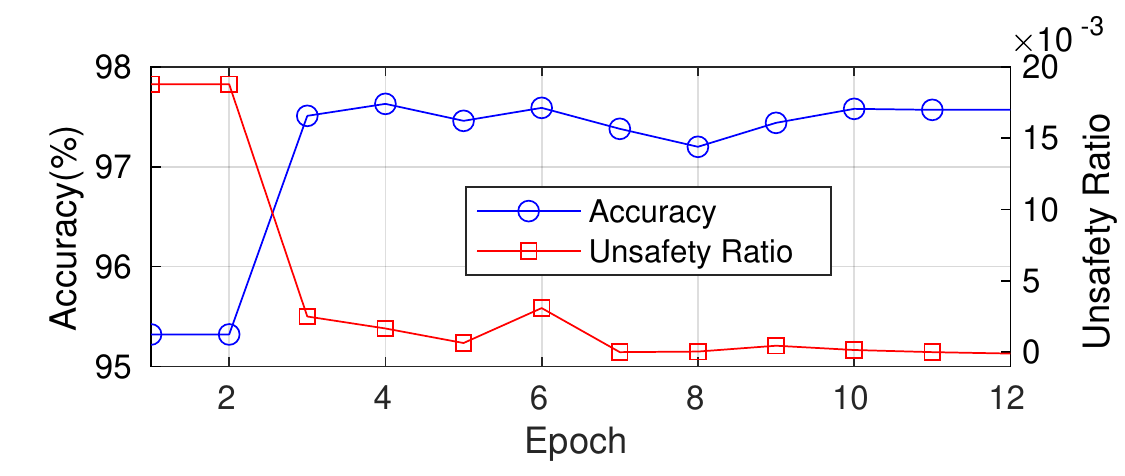}
    \end{subfigure}
    \begin{subfigure}[t]{0.42\textwidth}
        \centering
        \includegraphics[scale=0.5]{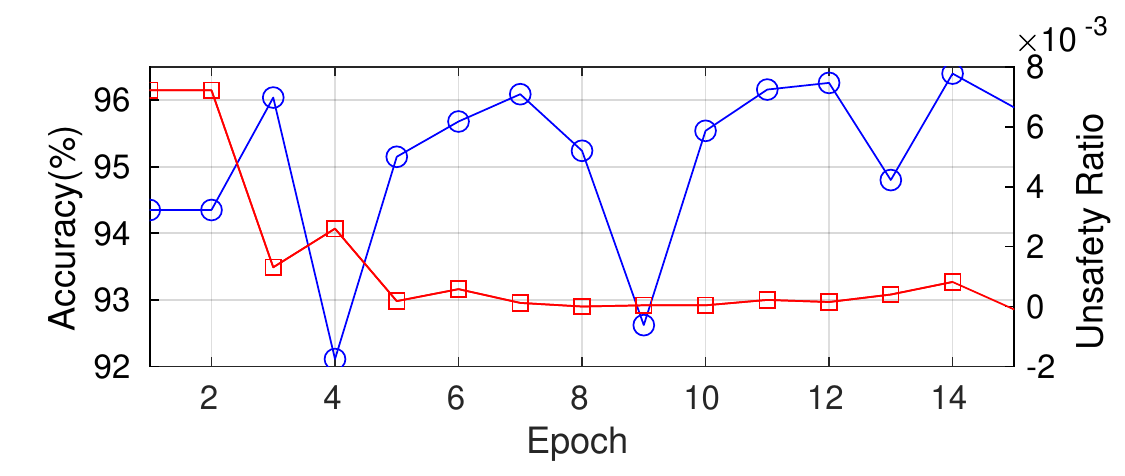}
    \end{subfigure}
    \caption{The evolution of the accuracy and the unsafe input domain in the repair of neural networks $N_{26}$ and $N_{17}$. The right $y$ axis represents the accuracy of the model candidate. The accuracy refers to the percentage of the correct classification of action advisory. The left $y$ axis represents the approximated volume ratio of the unsafe input domain to the whole input domain specified in the safety property.}
    \label{fig:HorizontalCAS_accuracy}
\end{figure}

\begin{table}[]
\caption{Comparison of our new reachability analysis method with the method~\cite{xiaodong21} on computational efficiency and memory efficiency. $\textbf{T}_r(sec)$ and $\textbf{M}_r(GB)$ denote the computational time and the maximum memory usage of our method for the reachability analysis in one repair. $\textbf{T}_{nr}(sec)$ and $\textbf{M}_{nr}(GB)$ are for~\cite{xiaodong21} on the same model candidate models.}
\resizebox{10cm}{!}{
\begin{tabular}{lccccccccccc} \toprule
Nets      & $N_{11}$ & $N_{12}$ & $N_{15}$ & $N_{16}$ & $N_{17}$ & $N_{19}$ & $N_{26}$ & $N_{41}$ & $N_{52}$ & $N_{55}$ & $N_{59}$ \\ \midrule
$\textbf{T}_r(sec)$  &3.3    &5.5    &4.6  & 5.2  & 61.6   & 9.0   &24.0   & 3.1    &3.1    &5.0    &3.3 \\
$\textbf{T}_{nr}(sec)$ & 3.5    &4.9    &5.7    &5.1   &56.8   & 9.0   &29.3   & 3.5   & 3.4   & 4.6   & 3.4     \\
$\textbf{M}_{r}(GB)$  &4.90    &4.91    &4.97    &4.92   & 4.97    &4.97    &4.96    &4.94    &4.94    &4.97    &4.94        \\
$\textbf{M}_{nr}(GB)$   &4.94    &4.97    &5.15    &4.99   & 5.23   & 5.19   & 5.12    &4.99    &5.00    &5.00    &5.01    \\ \bottomrule
\end{tabular}}
\label{tab:HorizontalCAS2}
\end{table}

More details of the repair process of $N_{26}$ and $N_{17}$ is included to illustrate the process. One is the evolution of the output reachable domain of the candidate model on the violated properties, as shown in Figure~\ref{fig:HorizontalCAS_sets}. We can notice that the reachable domain in blue expands with the repair while the unsafe domain in red gradually disappears. The expansion is mainly because the $\Delta \textbf{y}$ added to the unsafe $\textbf{y}$ changes the output distribution of the model candidate by turning away $\textbf{y}$ from the unsafe domain. The other one is the evolution of the accuracy and the unsafe input spaces of the candidate model, as shown in Figure~\ref{fig:HorizontalCAS_accuracy}. The unsafe input space computed for each property is quantified by the percentage of the unsafe volume related to the volume of the whole input domain. It can be approximated through a large amount of samplings. 
We can notice that the tendency of the ratio decreases along with the repair, indicating the unsafe input domain gradually disappears. While the new accuracy seldom goes below the original accuracy, indicating the repair in this cases does not degrade the performance.

Table~\ref{tab:HorizontalCAS2} describes the comparison of our new reachability analysis method to the method~\cite{xiaodong21} regarding computational and memory efficiency. We can notice that there are no obvious difference between the performance of these methods. This is because the computation of the reachability analysis is negligible so that the overhead can not be distinguished. 

\subsection{Rocket Lander Benchmark}

\begin{figure}[ht]
    \centering
    \includegraphics[scale=0.75]{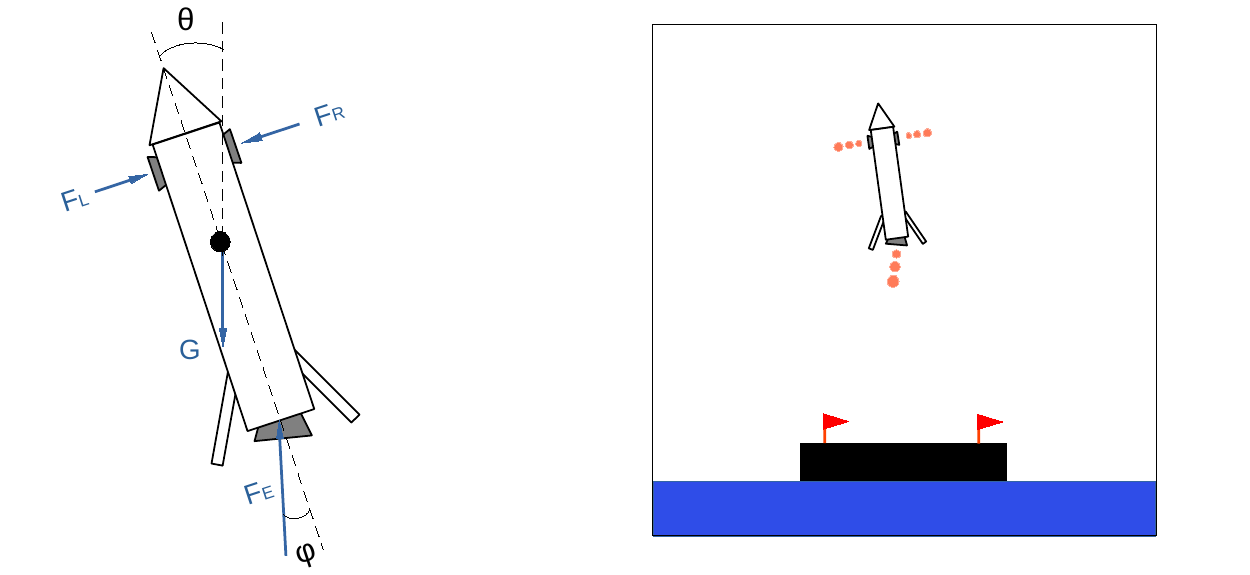}
    \caption{Rocket lander benchmark.}
    \label{fig:rocket-lander}
\end{figure}

The rocket lander benchmark is based on the lunar lander presented in~\cite{brockman2016openai}. It is a vertical rocket landing model simulating SpaceX's Falcon 9 first stage rocket. Unlike the lunar lander whose action space is discrete, the action space is continuous, which commonly exists in the practical applications. Besides the rocket, a barge is also included on the sea which moves horizontally and its dynamics are monitored. The benchmark is shown in Figure~\ref{fig:rocket-lander}. The rocket includes one main engine thruster at the bottom with an actuated joint and also two other side nitrogen thrusters attached to the sides of the top by un-actuated joints. The main engine has a power $F_E$ ranging in $[0,1]$ and its angle relative to the rocket body is $\varphi$. The power $F_S$ of the side thrusters ranges in $[-1, 1]$, where $-1$ indicates that the right thruster has full throttle and the left thruster is turned off while $1$ indicates the opposite. The rocket landing starts in certain height.  Its goal is to land on the center of the barge without falling or crashing by controlling its velocity and lateral angle $\theta$ through the thrusters. 


There are three actions, the main engine thruster $F_E$, its angle $\varphi$ and the side nitrogen thrusters $F_S$. The original observation contains the position $x$ and $y$ of the rocket relative to the landing center on the barge, the velocity $v_x$ and $v_y$ of the rocket, its lateral angle $\theta$, its angular velocity $\omega$. To improve the performance of agents, we also incorporate the last action advisory into the observation for reference. Then, the new observation can denoted as $[x, y, v_x, v_y, \theta, \omega, F_E', \varphi', F_S']$. Their lower bound $lb$ and upper bound $ub$ are in Equation~\ref{equ:lander-lbub}.
The starting state and the reward are similar to the lunar lander
The termination conditions of one \textit{episode} includes (1) $|x|>1$ which indicates the rocket moves out of the barge in $x$-space, (2) $y>1.3$ or $y<0$ which indicates the rocket moves out of the $y$-space or below the barge, (3) $\theta>35^{\circ}$ which indicates the rocket tilts greater than the controllable limit.
\begin{align}
\begin{split}
        lb &= [-\infty, 0, -\infty, -\infty, -\pi, -\infty, 0, -15^{\circ}, -1] \\
    ub &= [+\infty, +\infty, +\infty,+\infty, \pi, +\infty, 1, 15^{\circ}, 1]
\end{split}
\label{equ:lander-lbub}
\end{align}

Two safety properties are defined for the agent as below. Since reachability analysis processes bounded sets,
the infinite lower bounds and upper bounds of states above will be replaced with the searched state space in the learning process of the original agent.
\begin{enumerate}
    \item \textit{property 1}: for the state constraints $-20^{\circ}\leq\theta\leq-6^{\circ}$, $\omega<0$, $\varphi'\leq0^{\circ}$ and $F_S'\leq0$ , the desired action should be $\varphi< 0$ or $F_S< 0$, namely, the unsafe action domain is $\varphi\geq0 \cap F_S\geq 0$. It describes a scenario where the agent should always stop the rocket from tilting to the right. 
    \item \textit{property 2}: for the state constraints $6^{\circ}\leq\theta\leq20^{\circ}$, $\omega\geq0$, $\varphi'\geq0^{\circ}$ and $F_S'\geq0$ , the desired action should be $\varphi> 0$ or $F_S> 0$, namely, the unsafe action domain is $\varphi\leq0 \cap F_S\leq 0$. It describes a scenario where the agent should always stop the rocket from tilting to the left. 
\end{enumerate}

The reinforcement learning algorithm Deep Deterministic Policy Gradients (DDPG)~\cite{lillicrap2015continuous} is applied on this benchmark, which combines the Q-learning with Policy gradients. This algorithm is used for the environments with continuous action spaces. It consists of two models: Actor, a policy network that takes the state as input and outputs exact continuous actions rather than probability distribution over them, and Critic, a Q-value network that takes state and action as input and outputs Q-values. The Actor is our target agent controller with its safety properties.
DDPG uses experience replay to update Actor and Critic, where in the training process, a set of \textit{tuple}s are sampled from previous experiences.  

Here, we first learn several agents with the DDPG algorithm. Then, we apply our framework to repair agents that violate the safety properties. The architecture of the Actor is designed with 9 inputs for state, 5 hidden layers with each containing 20 ReLU neurons, 3 outputs with subsequent tanh function which maps input spaces into $[-1,1]$. Let the three outputs before the tanh be denoted as  $y_1$, $y_2$ and $y_3$, the outputs of Actor are computed by 
$F_E= 0.5\times\tanh(y_1)$+0.5, $\varphi=15^{\circ}\times\tanh(y_2)$ and $F_S=\tanh(y_3)$.
Our reachability analysis is applied to the architecture before the tanh function, which contains only ReLU neurons.
Although the unsafe output domains defined in safety properties above are for outputs $\varphi$ and $F_S$ after the tanh function, the domains $\varphi\geq0 \cap F_S\geq 0$, $\varphi\leq0 \cap F_S\leq 0$ are actually equivalent to $y_2\geq0 \cap y_3\geq 0$, $y_2\leq0 \cap y_3\leq 0$. The architecture of the Critic is designed with 12 inputs for state and action, 5 hidden layers with each containing 20 ReLU neurons, 1 output for the Q-value. The capacity of the global buffer to store previous experience is set to $4\times 10^5$.
The learning rate for Actor and Critic is set to $10^{-4}$ and $10^{-3}$ respectively. The 1000 \textit{episode}s are performed for each learning. Overall, three unsafe agents are obtained.

For the repair process, the parameters are as follows. the learning rates for Actor and Critic stay unchanged. A buffer stores all old experiences from the learning process. In addition, another global buffer is included to store new experiences with unsafe states as initial states. 
From these two buffers, old experiences as well as new experiences are randomly selected to form a set of training \textit{tuple}s in Figure~\ref{fig:framework-DRL}(b). As introduced, a new penalty reward is added for any wrong actions generated from input states. Its value is normally set to the lowest reward in the old experience. Here, the penalty is set to -30. To maintain the performance, the threshold of the change ratio which is defined in Equation~\ref{equ:ratio} is set to -0.2.
\begin{equation}
    \textit{Ratio} = \frac{ \textit{Performance}_{(repaired\ agent)}-\textit{Performance}_{( original\ agent)}}{\textit{Performance}_{(original\ agent)}}
    \label{equ:ratio}
\end{equation}

For each unsafe agent, we conduct the repair 5 times with each repair aiming to obtain a safe agent. There are 15 instances used for evaluation.  The experimental results are shown in Table~\ref{tab:lander1} and \ref{tab:lander2}. The evolution of the unsafe input domain like Figure~\ref{fig:HorizontalCAS_accuracy} is not included for this benchmark because the sampling does not work well for high dimensional input space. Table~\ref{tab:lander1} describes the performance change ratio, the epochs of repair and the total time, where the performance of agents is evaluated by the averaged reward on 1000 episodes of running. We note that our framework can successfully repair the 3 agents in all 15 instances. In most cases, the performance of the repaired agent is slightly improved. The performance degradation in other instances is also trivial. The repair process takes 2-6 epochs for all instances with the running time ranging from 332.7 seconds to 2632.9 seconds. Also, during the repair process, we notice that repairing unsafe behaviors on one property occasionally leads to new unsafe behaviors on the other property. This is likely because the input regions defined in the properties are adjacent to each other but their desired output regions are different, and the repaired behaviors over one input region can easily expand to other regions over which different behaviors are expected. 

Next, we conduct a comparison of our new reachability analysis method with the method presented in~\cite{xiaodong21}. The results are shown in Table~\ref{tab:lander2}. In terms of computational efficiency and memory efficiency, our method outperforms this method in~\cite{xiaodong21}. The computational efficiency improvement of our method ranges between a maximum of 6.8 times faster and a minimum of 3.6 times faster, with an average of 4.7. The reduction on memory usage ranges between 61.7\% and 70.2\%, with an average of f 64.5\%.  It is noteworthy that there are many factors that affect computational complexity of the reachability analysis, such as the number of neurons, the input domain as well as the parameter weights of DNNs. Therefore, for the three agents with the same architecture but different weights, the computational time is different.

\begin{table}[ht]
\caption{Repair of unsafe agents for the rocket lander. $\textbf{ID}$ is the index of each repair. $\textbf{Ratio}$ denotes the performance change ratio of the repaired agent compared to the original unsafe agent as formulated in Equation~\ref{equ:ratio}. $\textbf{Epoch}$ denotes the number of epochs for repair. $\textbf{Time}$ (\textit{sec}) denotes the running time for one repair with our reachability analysis method.
}
\resizebox{10cm}{!}{
\begin{tabular}{c|ccc|ccc|ccc} \toprule
             & \multicolumn{3}{c|}{Agent 1}                                                   & \multicolumn{3}{c|}{Agent 2}                                                   & \multicolumn{3}{c}{Agent 3}                                                   \\
 ID & $\textbf{Ratio}$ & $\textbf{Epoch}$ & $\textbf{Time}$  & $\textbf{Ratio}$  & $\textbf{Epoch}$ & $\textbf{Time}$ &  $\textbf{Ratio}$  & $\textbf{Epoch}$ & $\textbf{Time}$  \\ \midrule
1            & +0.063                 & 3      & 332.7                      & +0.048                  & 3      & 635.7                             & +0.053                  & 2      & 446.1                           \\
2            & +0.088                  & 3      & 302.0                    & +0.012                  & 6      & 1308.4                              & +0.085                  & 3      & 1451.6                         \\
3            & +0.079                  & 3      & 447.9                       & -0.084                  & 4      &  812.9                           & -0.033                  & 3      & 2417.1
                          \\
4            & +0.078                  & 3      & 884.2                      & +0.025                  & 3      & 620.3                              & +0.073                  & 2      & 1395.3                             \\
5            & +0.085                  & 3      & 754.3                       & -0.001                  & 4      & 813.5                           & -0.165                  & 5      & 2632.9                               \\  \bottomrule   
\end{tabular}}
\label{tab:lander1}
\end{table}

\begin{table}[ht]
\caption{Comparison of our new reachability analysis method with the method~\cite{xiaodong21} on computational efficiency and memory efficiency. $\textbf{T}_r(sec)$ and $\textbf{M}_r(GB)$ denote the computational time and the maximum memory usage of our method for all reachability analysis of all candidate models in one repair. $\textbf{T}_{nr}(sec)$ and $\textbf{M}_{nr}(GB)$ are for~\cite{xiaodong21} on the same model candidate models.
}
\resizebox{12cm}{!}{
\begin{tabular}{c|cc|cc|cc|cc|cc|cc} \toprule
             & \multicolumn{4}{c|}{Agent 1}                                                   & \multicolumn{4}{c|}{Agent 2}                                                   & \multicolumn{4}{c}{Agent 3}                                                   \\
 ID & $\textbf{T}_r$  & \multicolumn{1}{c}{$\textbf{T}_{nr}$}  & $\textbf{M}_r$ & $\textbf{M}_{nr}$  &  $\textbf{T}_r$ & \multicolumn{1}{c}{$\textbf{T}_{nr}$}  & $\textbf{M}_r$  & $\textbf{M}_{nr}$&  $\textbf{T}_r$& \multicolumn{1}{c}{$\textbf{T}_{nr}$} & $\textbf{M}_r$   & $\textbf{M}_{nr}$  \\ \midrule
1            &129.4    & 760.0     &15.94   & 42.69      & 426.1       & 1583.6    &13.38     & 36.39       & 700.5     & 2513.7   &14.48      & 46.05 \\
2            &122.9  & 740.4      &15.48      & 40.45      & 935.9     & 3352.8      &13.58    & 37.23        & 618.6       & 2586.6    &14.33     & 48.18 \\
3            &206.7     & 1361.9       &16.74      & 45.41      & 572.6      & 2106.7    &13.55     & 36.55        & 645.3     & 3054.0        &15.13   & 44.91 \\
4            &329.0        & 2250.6      &15.66     & 43.89      & 428.4      & 1569.7   &13.67      & 35.97        & 714.4      & 3019.8      &15.49   & 47.10 \\
5            &224.53     & 1454.2     &15.32     & 40.77       & 579.5      & 2108.3    &13.68     & 35.99       & 997.0        & 3277.3      &15.99   &48.86     \\  \bottomrule   
                
\end{tabular}}
\label{tab:lander2}
\end{table}

\begin{figure}[ht]
    \centering
    \begin{subfigure}[t]{0.27\textwidth}
        \centering
        \includegraphics[scale=0.28]{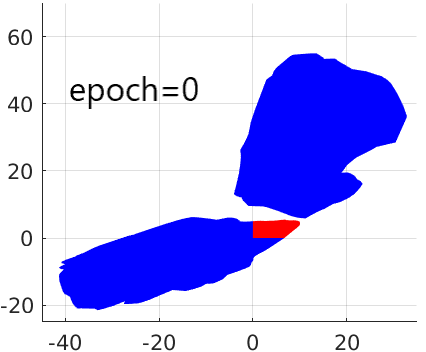}
    \end{subfigure}
    \begin{subfigure}[t]{0.27\textwidth}
        \centering
        \includegraphics[scale=0.28]{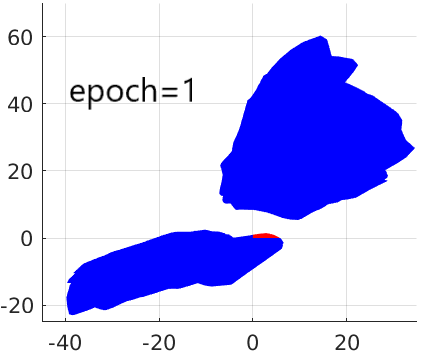}
    \end{subfigure}
    ~
    \begin{subfigure}[t]{0.27\textwidth}
        \centering
        \includegraphics[scale=0.28]{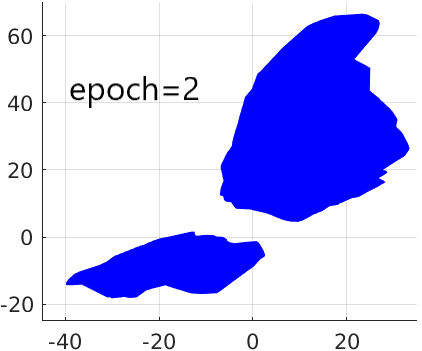}
    \end{subfigure}
    \caption{The evolution of the output reachable domain and the unsafe reachable domain in the repair of Agent 1 on the first repair. $x$ axis represents $y_2$ and $y$ axis represents $y_3$. The blue area represents the exact output reachable domain while the red area represents the unsafe reachable domain which is a subset of the exact output reachable domain. The bottom left area is the reachable domain on Property 1 and the top right area is the reachable domain on Property 2.}
    \label{fig:lander_sets}
\end{figure}

In addition, we analyze the evolution of the reachable domain of candidate models in the repair. An example of Agent 1 on the first repair is shown in Figure~\ref{fig:lander_sets}. At $epoch=0$ which is before the repair, we can notice that the candidate agent has unsafe output reachable domain on Property 1 and is safe on Property 2. At $epoch=1$, the unsafe reachable domain becomes smaller, which indicates that the agent learned from the penalty assigned to the unsafe actions and its action space moves towards the safe domain. After another repair, the agent becomes safe on both properties. We can also notice that during the repair the output reachable domains do not change much, indicating that the performance of the agent is preserved. 

\section{Related Works}

\textbf{Adversarial and robust training.}  The adversarial training~\cite{goodfellow2014explaining, madry2017towards,wong2018provable, mirman2018differentiable, zhang2019theoretically} is a type of method where adversarial examples are obtained by adversarial attacks or reachability analysis based on the over approximation. These methods have been shown effective in improving the robustness of DNN against adversarial attacks. It inspires us to combine the adversarial training with the more accurate reachability analysis methods that provide complete details of misbehaviors, such that provably safe DNNs can be learned. The difference between our framework and the adversarial training is that instead of examples from random attacks, our framework can identify examples representative of the entire unsafe domain computed from the reachability analysis.

\textbf{Verification of Deep Neural Networks.} Many methods for safety verification of DNNs have been developed, which are mainly based on \textit{reachability}~\cite{gehr2018ai2,xiang2018output,tran2019star, yang2020reachability, xiang2017reachable}, \textit{optimization}~\cite{bastani2016measuring, lomuscio2017approach, raghunathan2018certified, dvijotham2018dual, tjeng2018evaluating, wong2018provable}, and \textit{search}~\cite{wang2018formal,huang2017safety,weng2018towards,katz2017reluplex, ehlers2017formal, bunel2018unified, dutta2018output}. The reachability analysis method is a very appealing method for the repair because it can provide regions of DNN misbehaviors.
The spectrum of reachability methods can be broadly categorized in two classes: over-approximation and exact analysis methods. Over-approximation methods are able to provide safety guarantees but are also \textit{incomplete}, i.e., the method may return \textit{unsafe} due to over-approximation when in fact the system is \textit{safe}. These methods ensure quick safety verification but are not sufficient for the repair. The repair also requires the exact analysis method which can compute the exact unsafe input domain and output domain of DNNs. Thus, we design an algorithm to integrate these two type of method by taking advantage of their merits, which has been shown around 5 times faster than the related work.

The efficiency and accuracy of reachability analysis is strongly associated with the set representation. Approaches, particularly for the 
DNN's, \textit{Zonotope}~\cite{gehr2018ai2}, \textit{Star-set}~\cite{tran2019fm,tran2020nnv, tran2021cav, tran2020cav, bak2020cav, tran2019emsoft} and \textit{facet-vertex incidence matrix} (FVIM) with vertices~\cite{xiaodong21} are utilized.
Each set representation has its advantages and challenges. For instance, a \textit{zonotope} can be represented with finite vectors $\textbf{v}_i$ by summing $a_i\textbf{v}_i$, where $a_i$ is scalar ranging between 0 and 1. Such a simple representation enables the development of fast over-approximation methods for reachability analysis. The \textit{star-set} representation is essentially an enhancement of the \textit{half-space representation}, which can efficiently process affine mapping in DNNs. The FVIM is used for exact reachability analysis for repair in this work. For the integration of methods, the new set representation $\mathcal{V}$-zono which can be compatible with FVIM and efficiently encodes vertices is designed for the over approximation method with ReLU linear relaxation. 

\textbf{Deep Neural Networks Repair.} Works~\cite{sohn2019search, goldberger2020minimal} attempt to correct unsafe behavior of DNNs by modifying neural weights that is likely associated with the misbehaviors. Due to the black-box nature of DNNs, the modification of such weights may result in unpredicted performance degradation of DNNs. Work~\cite{sotoudeh2021provable} introduces a \textit{Decoupled} DNN architecture. Based on this architecture, their provable polytope repair which aims to correct misbehaviors of ReLU DNNs over a domain can be reduced to a LP problem.  However, this method is only applicable to the two-dimensional input region for the DNNs having similar size to the ones in the HorizontalCAS benchmark -- a five-dimensional input.

\section{Conclusion and Future Work}
We have presented a reachability-based framework to repair unsafe DNN 
controllers for autonomous systems. 
The approach can be utilized to repair unsafe DNNs with only training data available.  It can also be integrated into existing reinforcement algorithms to synthesize safe DNN controllers. 
Our experimental results on two practical benchmarks have shown that the proposed framework can successfully obtain a provably safe DNN while maintaining its accuracy and performance. We utilize a new set representation and integrate an over approximation method to improve the performance and memory footprint of our rechability analysis algorithm. 

Nonetheless, safe training or repairing of DNNs with reachability analysis is still a challenging problem. There are several aspects we plan to study in the future. Firstly, computation of the unsafe set domain of larger-scale DNNs with higher dimensional inputs, such as DNNs for image classification, is still challenging. Therefore, new approaches are needed to repair such unsafe DNNs. Secondly, training of DNNs relies on appropriate meta parameters and cannot always guarantee the convergence to optimal performance, which can impose difficulties for the repair process to converge. Thus, analysis of the interaction between DNN training and its repair is necessary.
Furthermore, from our observations it becomes more difficult to repair unsafe DNNs when the input spaces of two safety properties are adjacent but with different desired output behaviors. This is due to the fact that it may be difficult for the DNN to learn a boundary to distinguish these adjacent input spaces and behave correctly. Therefore, a thorough study on understanding the convergence of the proposed framework is critical for enhancing its applicability to real-world applications.

\newpage
\bibliography{bibfile.bib, bibfile2.bib}

\end{document}